\documentclass[10pt, twocolumn, letterpaper, table]{article}

\usepackage[pagenumbers]{cvpr} 

\usepackage{multirow}
\usepackage{makecell}
\usepackage{listings}
\usepackage{xcolor}

\definecolor{cvprblue}{rgb}{0.21,0.49,0.74}
\usepackage[pagebackref,breaklinks,colorlinks,allcolors=cvprblue]{hyperref}

\def\paperID{xxxx} 
\def\confName{CVPR}
\def\confYear{2025}

\newcommand{\upfast}[1]{\tiny\textcolor{red}{#1}}

\title{CLEAR: Conv-Like Linearization Revs Pre-Trained Diffusion Transformers Up}

\author{Songhua Liu \quad Zhenxiong Tan \quad Xinchao Wang$^*$\\
National University of Singapore\\
{\tt\small \{songhua.liu, zhenxiong\}@u.nus.edu, xinchao@nus.edu.sg}
}

\begin{document}

\twocolumn[{%
\renewcommand\twocolumn[1][]{#1}%
\maketitle
\begin{center}
    \centering
    \vspace{-0.6cm}
    \captionsetup{type=figure}
    \includegraphics[width=\textwidth]{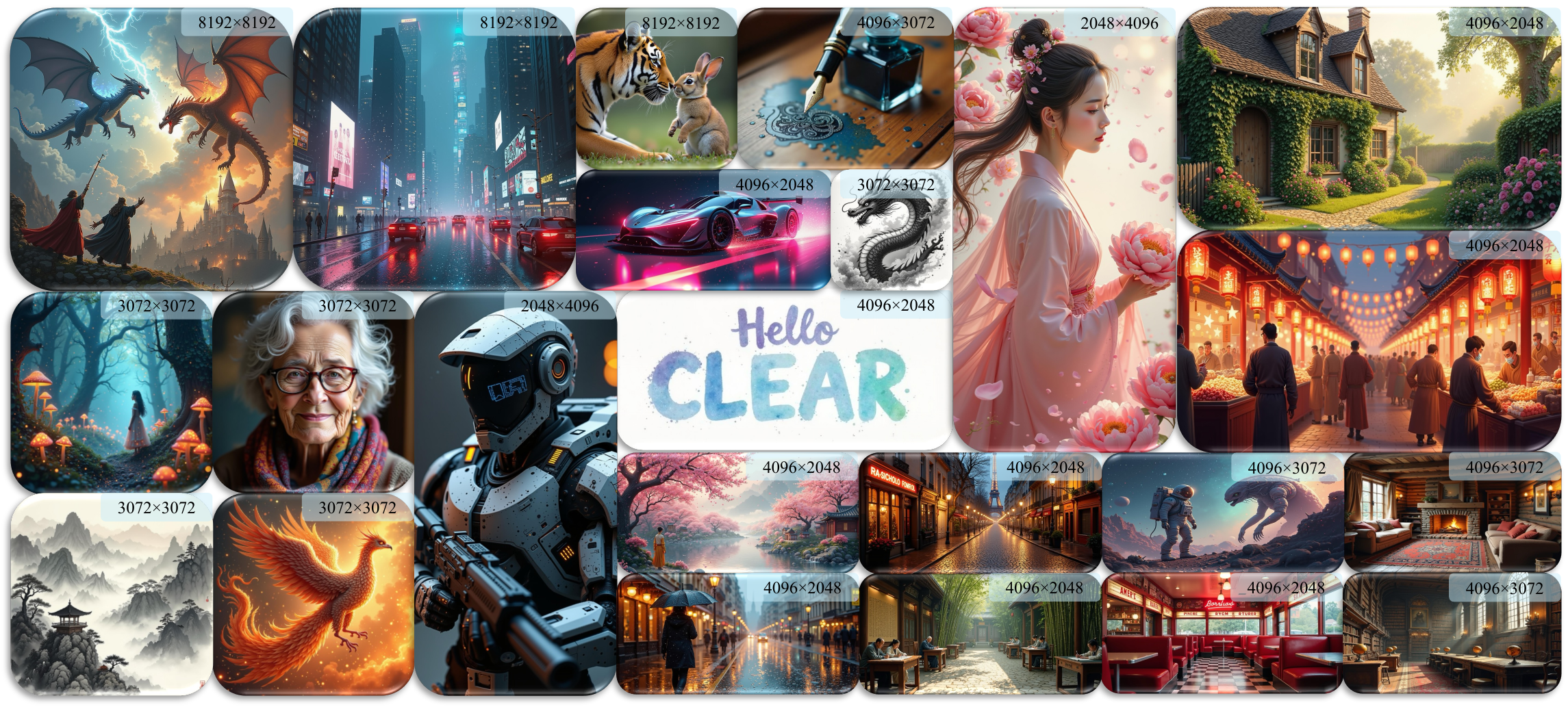}
    \vspace{-0.8cm}
    \caption{Ultra-resolution results generated by the linearized FLUX.1-dev model with our approach CLEAR. Resolution is marked on the top-right corner of each result in the format of \texttt{width}$\times$\texttt{height}. Corresponding prompts can be found in the appendix. 
    \label{fig:teaser}
    } 
\end{center}%
}]

\renewcommand{\thefootnote}{\fnsymbol{footnote}}
\footnotetext[1]{Corresponding Author.}
\renewcommand{\thefootnote}{\arabic{footnote}}

\begin{abstract}
Diffusion Transformers (DiT) have become a leading architecture in image generation.  
However, the quadratic complexity of attention mechanisms, which are responsible for modeling token-wise relationships, results in significant latency when generating high-resolution images. 
To address this issue, we aim at a linear attention mechanism in this paper that reduces the complexity of pre-trained DiTs to linear. 
We begin our exploration with a comprehensive summary of existing efficient attention mechanisms and identify four key factors crucial for successful linearization of pre-trained DiTs: locality, formulation consistency, high-rank attention maps, and feature integrity. 
Based on these insights, we introduce a convolution-like local attention strategy termed CLEAR, which limits feature interactions to a local window around each query token, and thus achieves linear complexity. 
Our experiments indicate that, by fine-tuning the attention layer on merely 10K self-generated samples for 10K iterations, we can effectively transfer knowledge from a pre-trained DiT to a student model with linear complexity, yielding results comparable to the teacher model. 
Simultaneously, it reduces attention computations by 99.5\% and accelerates generation by 6.3 times for generating 8K-resolution images. 
Furthermore, we investigate favorable properties in the distilled attention layers, such as zero-shot generalization cross various models and plugins, and improved support for multi-GPU parallel inference. 
Models and codes are available \href{https://github.com/Huage001/CLEAR}{here}. 
\end{abstract}    
\section{Introduction}
\label{sec:intro}

Diffusion models~\cite{nichol2021improved,dhariwal2021diffusion,rombach2022high,ho2020denoising} have gained widespread attention in text-to-image generation, proving to be highly effective for producing high-quality and diverse images from textual prompts~\cite{croitoru2023diffusion,yang2023diffusion}. 
Traditionally, architectures based on UNet~\cite{ronneberger2015u,rombach2022high} have dominated this field due to their robust generative capabilities. 
In recent years, Diffusion Transformers (DiTs)~\cite{peebles2023scalable,bao2023all,chen2023pixart,esser2024scaling,li2024hunyuan,gao2024lumina,chen2024pixartsigma} have emerged as a promising alternative, achieving leading performance in this field. 
Unlike the UNet-based architectures, DiTs leverage the attention mechanism~\cite{vaswani2017attention} to model intricate token-wise relationships with remarkable flexibility, enabling them to capture nuanced dependencies across all tokens in images and texts, and thus produce visually rich and coherent outputs. 

\begin{figure}[t]
  \centering
   \includegraphics[width=\linewidth]{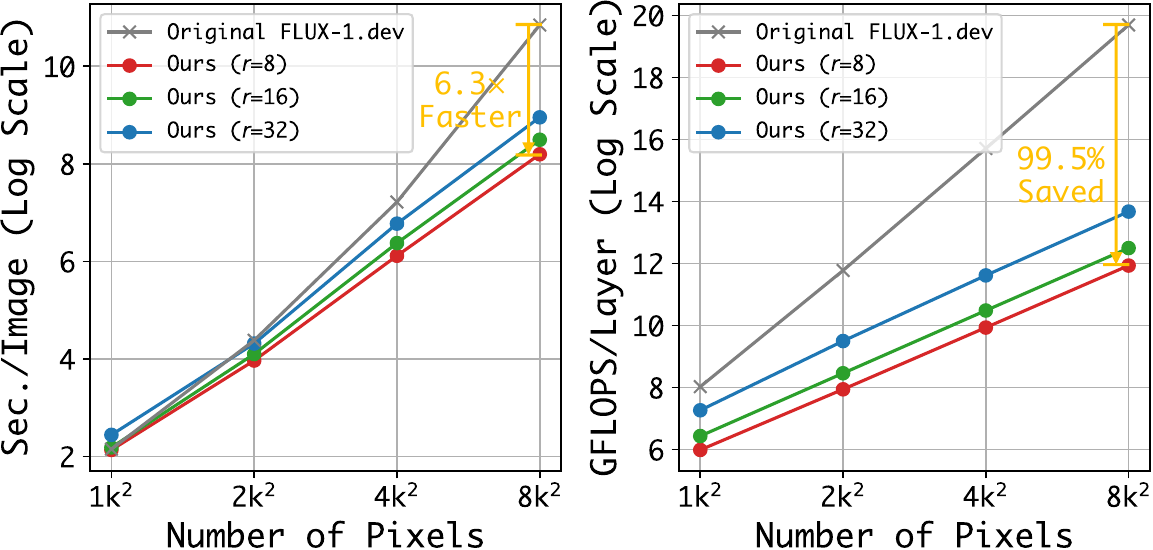}
   \vspace{-0.7cm}
   \caption{Comparison of speed and GFLOPS between the proposed linearized DiT and the original FLUX.1-dev. Speed is evaluated by performing 20 denoising steps on a single H100 GPU. FLOPS is calculated with the approximation: $4\times\sum M\times c$, where $c$ is the feature dimension and $M$ denotes the attention masks. $\log_2$ is applied on both vertical axes for better visualization. The raw data are supplemented in the appendix.}
   \vspace{-0.5cm}
   \label{fig:2}
\end{figure}

Despite their impressive performance, the attention layers—which model intricate pairwise token relationships with quadratic complexity—can introduce substantial latency in high-resolution image generation. 
As shown in Fig.~\ref{fig:2}, FLUX.1-dev~\cite{blackforestlabs_flux}, a state-of-the-art text-to-image DiT, requires over 30 minutes to generate 8K-resolution images with 20 denoising steps, even with hardware-aware optimizations like FlashAttention~\cite{dao2022flashattention,dao2023flashattention}. 

Focusing on these drawbacks, we are curious about one question in this paper: \emph{Is it possible to convert a pre-trained DiT to achieve linear complexity?} 
The answer is not straightforward, in fact, as it remains unclear whether existing efficient attention mechanisms—despite their recent widespread exploration~\cite{dao2024transformers,katharopoulos2020transformers,yang2023gated,ramapuram2024theory,wang2020linformer,roy2021efficient,beltagy2020longformer,zaheer2020big,choromanski2020rethinking,you2024linear,zhang2024gated,hua2022transformer,han2025agent}—can be effectively applied to pre-trained DiTs. 

To answer this question, we initiate our exploration with a summary of previous methods dedicated to efficient attention, categorizing them into three main strategies: formulation variation, key-value compression, and key-value sampling. 
We then experiment with fine-tuning the model by replacing the original attention layers with these efficient alternatives. 
Results indicate that while formulation variation strategies have proven effective in attention-based UNets~\cite{liu2024linfusion} and DiTs trained from scratch~\cite{xie2024sana}, they do not yield similar success with pre-trained DiTs. 
Key-value compression often leads to distorted details, and key-value sampling highlights the necessity of local tokens for each query to generate visually coherent results. 

Building on these observations, we figure out four elements crucial for for linearizing pre-trained DiTs, including locality, formulation consistency, high-rank attention maps, and feature integrity. 
Satisfying all these criteria, we present a \underline{c}onvolution-like \underline{l}in\underline{ear}ization strategy termed \emph{CLEAR}, where each query interacts only with tokens within a predefined distance $r$. 
Since the number of key-value tokens interacting with each query is fixed, the resulting DiT achieves linear complexity with respect to image resolution. 

To our surprise, such a concise design yields results comparable to original FLUX.1-dev after a knowledge distillation process~\cite{hinton2015distilling} with merely 10K fine-tuning iterations on 10K self-generated samples. 
As shown in Fig.~\ref{fig:teaser}, CLEAR exhibits satisfactory cross-resolution generalizability, a property also reflected in UNet-based diffusion models~\cite{bar2023multidiffusion,du2024demofusion,he2023scalecrafter,huang2024fouriscale}. 
For ultra-high-resolution generation like 8K, it reduces attention computations by 99.5\% and accelerates the original DiT by 6.3 times, as shown in Fig.~\ref{fig:2}. 
The distilled local attention layers are also compatible with different variants of the teacher model, \textit{e.g.}, FLUX.1-dev and FLUX.1-schnell, and various pre-trained plugins like ControlNet~\cite{zhang2023adding} without requiring any adaptation. 

As the token interactions are performed locally, it is convenient for CLEAR to support multi-GPU parallel inference. 
We further develop a patch-parallel paradigm that minimizes communication overhead. 
Our contribution can be summarized as follows:
\begin{itemize}
    \item We provide a taxonomic overview of recent efficient attention mechanisms and identity four elements essential for linearizing pre-trained DiTs. 
    \item Based on them, we propose a convolution-like local attention mechanism termed CLEAR as an alternative to default attention, which is the first linearization strategy tailored for pre-trained DiT to the best of our knowledge. 
    \item We delve into multiple satisfactory properties of CLEAR through experiments, including its comparable performance with the original DiT, linear complexity, cross-resolution generalizability, cross-model/plugin generalizability, support for multi-GPU parallel inference, \textit{etc.} 
\end{itemize}
The rest of this paper is organized as follows: we summarize recent efficient attention mechanisms in Sec.~\ref{sec:2}, present our main method in Sec.~\ref{sec:3}, experimentally validate its effectiveness in Sec.~\ref{sec:4}, and finally concludes this paper in Sec.~\ref{sec:5}. 

\section{Efficient Attention: A Taxonomic Overview}
\label{sec:2}

The attention mechanism~\cite{vaswani2017attention} is known for its flexibility in modeling token-wise relationships. 
It takes a query matrix $Q\in\mathbb{R}^{n\times c}$, a key matrix $K\in\mathbb{R}^{m\times c}$, and a value matrix $V\in\mathbb{R}^{m\times c'}$ as input and produces an output matrix $O\in\mathbb{R}^{n\times c'}$ via:
\begin{equation}
    O=\mathrm{softmax}(\frac{QK^\top}{\sqrt{c}})V,\label{eq:1}
\end{equation}
where $n$ and $m$ are the numbers of query and key-values tokens respectively, and $c$ and $c'$ are the feature dimensions for query-key and value tokens. 
In line with standard design conventions, we assume $c=c'$ throughout this paper, and in the case of self-attention, $Q$, $K$, and $V$ come from the same feature maps with $m=n$. 

As shown in Eq.~\ref{eq:1}, self-attention involves constructing $n\times n$ attention maps to model pair-wise token-to-token relationships, which results in both time and memory complexity. 
To address this issue, numerous studies focus on developing efficient attention mechanisms. In this section, we summarize recent work in this area and assess its applicability to DiT linearization. Specifically, we categorize existing approaches into three main categories: formulation variation, key-value compression, and key-value sampling. 

\subsection{Formulation Variation}

Revisiting Eq.~\ref{eq:1}, if the \texttt{softmax} operation is omitted, we can first compute $K^\top V$, yielding a $c\times c$ matrix with linear time in relation to $n$. 
In this way, a series of linear attention mechanisms apply kernel functions $f(\cdot)$ and $g(\cdot)$ to $Q$ and $K$ respectively to mimic the effect of \texttt{softmax}:
\begin{equation}
    O=f(Q)g(K)^\top V,
\end{equation}
such as Mamba2~\cite{dao2024transformers}, Gated Linear Attention~\cite{yang2023gated}, and Generalized Linear Attention~\cite{liu2024linfusion}. 
Another mainstream of methods try to replace the \texttt{softmax} operation with efficient alternatives, like \texttt{sigmoid}~\cite{ramapuram2024theory}, \texttt{$\mathrm{relu^2}$}~\cite{hua2022transformer,you2024linear}, and Nystrom-based approximation~\cite{xiong2021nystromformer}. 

\subsection{Key-Value Compression}

In the default setting of self-attention, the numbers of query and key-value tokens are consistent, \textit{i.e.}, $m=n$, and the shape of the attention map would be $n\times n$. 
It is thus promising to compress key-value tokens so that $m$ can be smaller than $n$ to reduce the complexity. 
Following this routine, PixArt-Sigma~\cite{chen2024pixartsigma} compress KV tokens locally with a downsampling \texttt{Conv2d} operator. 
Agent Attention~\cite{han2025agent} first conducts attention with downsampled $Q$ and full-sized $K$ and $V$ to select agent KV tokens for compression. 
Then, original $Q$ would interact with these compressed tokens. 
Similarly, Slot Attention~\cite{zhang2024gated} adopts learnable slots to obtain agent KV. 
Linformer~\cite{wang2020linformer} introduces learnable maps to obtain compressed tokens from the original ones. 

\subsection{Key-Value Sampling}

Efficient attention based on key-value sampling is based on the assumption that not all key-value tokens are important for a query and the attention matrix is highly sparse. 
Comparing with key-value compression, it prunes original key-value tokens for each token instead of producing new key-value tokens. 
For instance, Strided Attention~\cite{child2019generating} samples one key-value token at a regular interval. 
Routing Attention~\cite{roy2021efficient} samples key-value tokens based on grouping. 
Swin Transformer~\cite{liu2021swin} divides feature maps into non-overlapping local windows and performs attention independently for each window. 
Neighborhood Attention~\cite{hassani2023neighborhood} selects key-value tokens within a local window around each query. 
BigBird~\cite{zaheer2020big} uses a token selection strategy combining neighborhood attention and random attention, and LongFormer~\cite{beltagy2020longformer} combines neighborhood attention with some global tokens that are visible to all tokens. 
\section{Methods}
\label{sec:3}

\begin{figure*}[!t]
  \centering
  \includegraphics[width=\linewidth]{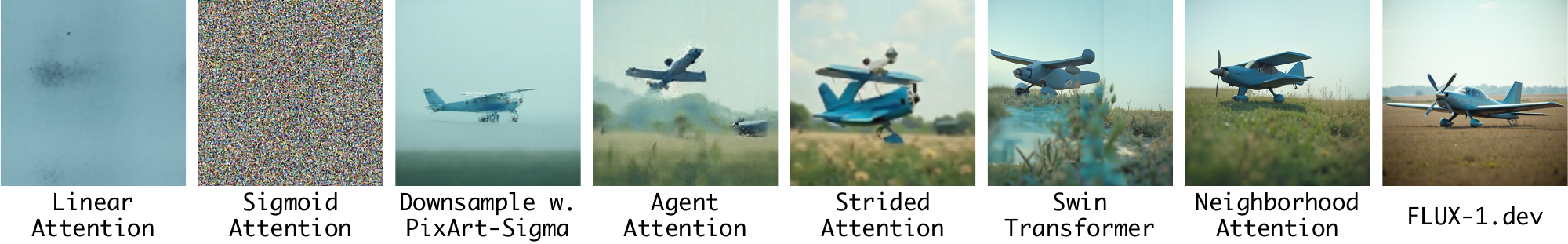}
  \vspace{-0.7cm}
  \caption{Preliminary results of various efficient attention methods on FLUX-1.dev. The prompt is ``\texttt{A small blue plane sitting on top of a field}".}
  \vspace{-0.6cm}
  \label{fig:3}
\end{figure*}

\subsection{What are Crucial for Linearizing DiTs?}\label{sec:3-1}

\begin{table}[!t]
    \centering
    \resizebox{\linewidth}{!}{
    \begin{tabular}{ccccc}
\toprule
\textbf{Method}                 & \textbf{Locality} & \begin{tabular}[c]{@{}c@{}}\textbf{Formulation}\\ \textbf{Consistency}\end{tabular} & \begin{tabular}[c]{@{}c@{}}\textbf{High-Rank}\\ \textbf{Attention Maps}\end{tabular} & \begin{tabular}[c]{@{}c@{}}\textbf{Feature}\\ \textbf{Integrity}\end{tabular} \\
\midrule
Linear Attention~\cite{dao2024transformers,liu2024linfusion,yang2023gated,katharopoulos2020transformers}                 & Yes      & No                                                                & No                                                                 & Yes                                                         \\
Sigmoid Attention~\cite{ramapuram2024theory}      & Yes      & No                                                                & Yes                                                                & Yes                                                         \\
PixArt-Sigma~\cite{chen2024pixartsigma}           & Yes      & Yes                                                               & Yes                                                                & No                                                          \\
Agent Attention~\cite{han2025agent}        & Maybe    & Yes                                                               & Yes                                                                & No                                                          \\
Strided Attention~\cite{child2019generating}      & No       & Yes                                                               & Yes                                                                & Yes                                                         \\
Swin Transformer~\cite{liu2021swin}       & Yes      & Yes                                                               & No                                                                 & Yes                                                         \\
Neighborhood Attention~\cite{hassani2023neighborhood} & Yes      & Yes                                                               & Yes                                                                & Yes                                                         \\
\bottomrule
\end{tabular}
    }
    \vspace{-0.3cm}
    \caption{Summary of existing efficient attention mechanisms based on the four factors crucial for linearizing DiTs.}
    \vspace{-0.5cm}
    \label{tab:sum}
\end{table}

Building on the overview of recent efficient attention mechanisms in Sec.~\ref{sec:2}, we explore a key question here: \emph{What specific features are essential for successfully linearizing pre-trained DiTs}? 
We thus try substituting all the attention layers in FLUX.1-dev with various efficient alternatives and fine-tuning parameters in these layers. 
The preliminary text-to-image results are shown in Fig.~\ref{fig:3}, through which we figure out four key elements: locality, formulation consistency, high-rank attention maps, and feature integrity. 
According to these perspectives, we summarize some previous efficient attention methods in Tab.~\ref{tab:sum}. 

\textbf{Locality} indicates that key-value tokens fallen in the neighborhood of a query are included for attention.  
From Fig.~\ref{fig:3}, we observe that many methods equipped with this feature yield at least plausible results, like PixArt-Sigma, Swin Transformer, and Neighborhood Attention. 
Particularly, comparing the results of Neighborhood Attention and Strided Attention, we find that incorporating local key-value tokens diminishes a lot of distorted patterns. 

\begin{figure}[t]
  \centering
   \includegraphics[width=\linewidth]{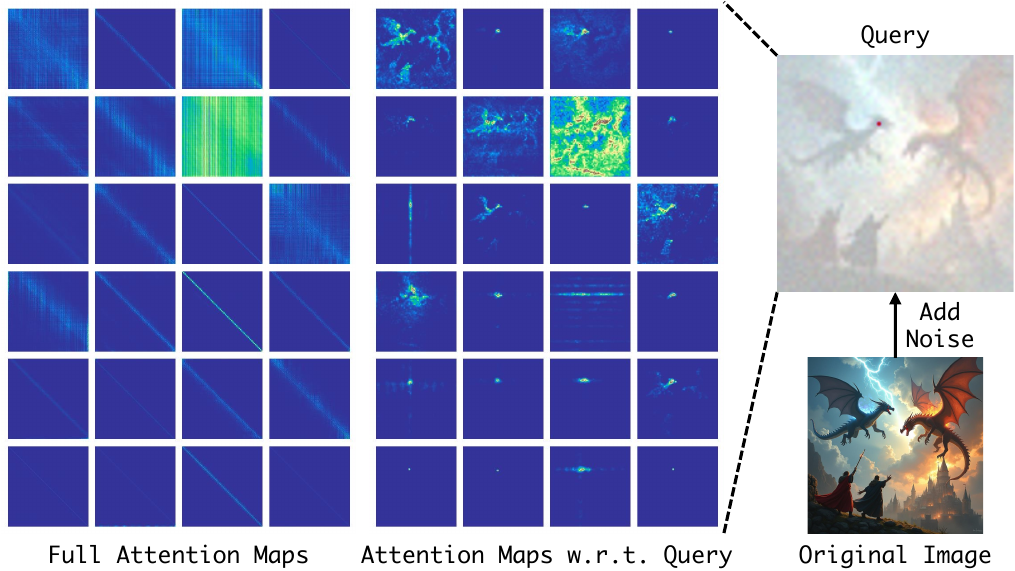}
   \vspace{-0.7cm}
   \caption{Visualization of attention maps by various heads for an intermediate denoising step. Attention in pre-trained DiTs is largely conducted in a local fashion.}
   \vspace{-0.7cm}
   \label{fig:4}
\end{figure}

The reason for these phenomenons is that pre-trained DiTs, such as FLUX, rely heavily on local features to manage token relationships. 
To validate this, we visualize attention maps in Fig.~\ref{fig:4} and observe that most significant attention scores fall in the local area around each query. 

\begin{figure}[t]
  \centering
   \includegraphics[width=\linewidth]{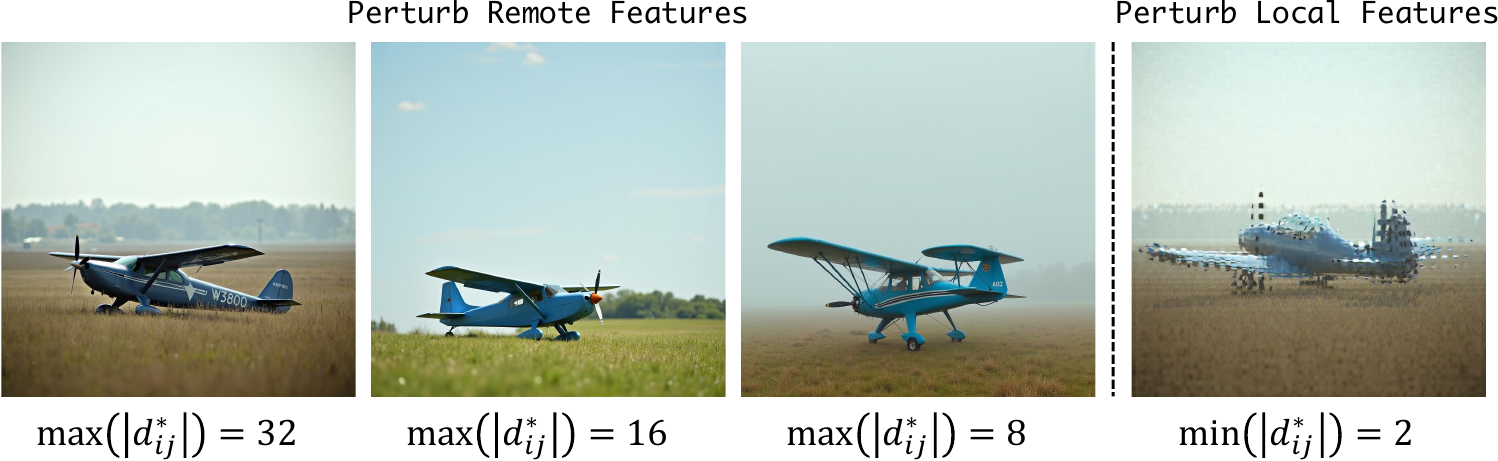}
   \vspace{-0.7cm}
   \caption{We try perturbing remote and local features respectively through clipping the relative distances required for rotary position embedding. Perturbing remote features has no obvious impact on image quality, whereas altering local features results in significant distortion. The text prompt and the original generation result are consistent with Fig.~\ref{fig:3}.}
   \vspace{-0.7cm}
   \label{fig:5}
\end{figure}

And in Fig.~\ref{fig:5}, we provide further evidence to illustrate the importance of local features, that perturbing remote features would not damage the quality of FLUX.1-dev much. 
Specifically, FLUX.1-dev relies on rotary position embedding~\cite{su2024roformer} to perceive spatial relationships and is sensitive to the relative distance $(d_{ij}^{(x)},d_{ij}^{(y)})$ on the two axes of a 2D feature map, where indices $i$ and $j$ denotes query and key token indices respectively. 
We perturb remote features by clipping the relative distances for rotary position embedding to a maximum value $r$ when they exceed this threshold, \textit{i.e.}, $d_{ij}^{(*)'}=d_{ij}^{(*)}\mathrm{.clip}(-r, r)$. 
As shown in Fig.~\ref{fig:5}(left), the results are reasonable for a $64\times64$ feature map when $r$ is as small as $8$. 
Conversely, if we perturb local features by setting their minimum absolute distances to $r$, even with $r$ as small as $2$, the result still collapses as shown in Fig.~\ref{fig:5}(right)—emphasizing the importance of locality. 

\textbf{Formulation Consistency} denotes that the efficient attention still apply the \texttt{softmax}-based formulation of the scaled dot-product attention. 
LinFusion~\cite{liu2024linfusion} has shown that linear attention approaches like linear attention achieve promising results in attention-based UNets. 
However, we find that it is not the case for pre-trained DiTs, as shown in Fig.~\ref{fig:3}. 
We speculate that it is due to attention layers being the only modules for token interactions in DiTs, unlike the case in U-Nets. 
Substituting all of them would have a substantial impact on the final outputs. 
Other formulations like Sigmoid Attention fails to converge within a limited number of iterations, unable to mitigate the divergence between the original and modified formulations. 
It is thus beneficial to maintain consistency with the original attention function. 

\textbf{High-Rank Attention Maps} means that attention maps calculated by efficient attention alternatives should be sufficient to capture the intricate token-wise relationships. 
As visualized in Fig.~\ref{fig:4}, extensive attention scores are concentrated along the diagonal, indicating that the attention maps do not exhibit the low-rank property assumed by many prior works. 
That is why methods like linear attention and Swin Transformer largely produce blocky patterns. 

\textbf{Feature Integrity} implies that raw query, key, and value features are more favorable than the compressed ones. 
Although PixArt-Sigma has demonstrated that applying KV compression on deep layers would not hurt the performance much, this approach is not suitable for completely linearizing pre-trained DiTs. 
As shown in Fig.~\ref{fig:3}, methods based on KV compression, such as PixArt-Sigma and Agent Attention, tend to produce distorted textures compared to the results from Swin Transformer and Neighborhood Attention, which highlights the necessity to preserve the integrity of the raw query, key, and value tokens. 

\subsection{Conv-Like Linearization}

Given the above analysis of the crucial factors for linearizing DiTs, Neighborhood Attention is the only scheme satisfying all the four constraints. 
Motivated on this, we propose CLEAR, a \underline{c}onv-like \underline{l}in\underline{ear}ization strategy tailored for pre-trained DiTs. 
Specifically, given that state-of-the-art DiTs for text-to-image generation, like FLUX and StableDiffusion 3 series~\cite{esser2024scaling}, typically adopt text-image joint self-attention for feature interaction, for each text query, it still gathers features from all text and image key-value tokens; while for each image query, it interacts with all text tokens and local key-value tokens fallen in a local window around it. 
Since the number of text tokens and the local window size remain constant as resolution increases, the overall complexity scales linearly with the number of image tokens. 

\begin{figure}[t]
  \centering
   \includegraphics[width=\linewidth]{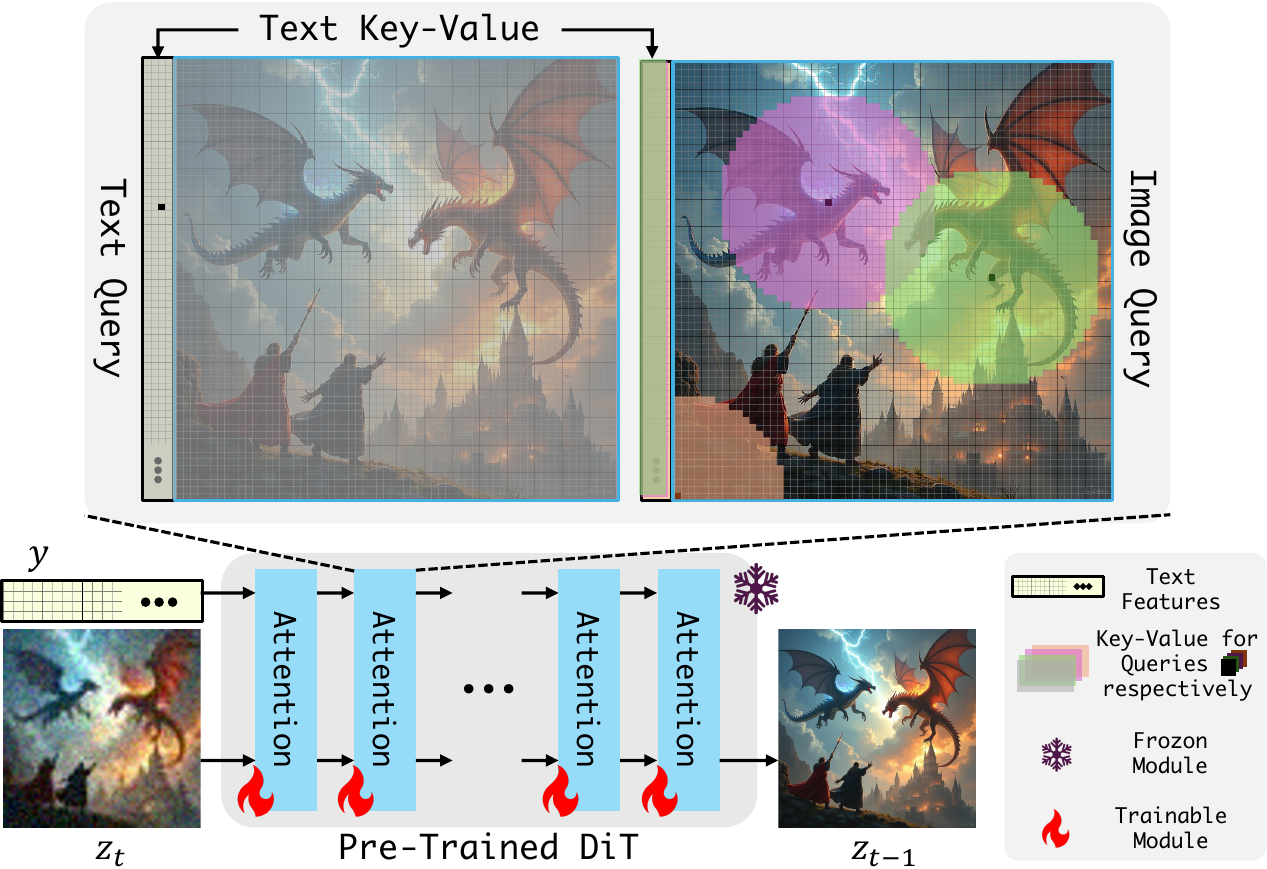}
   \vspace{-0.7cm}
   \caption{Illustration of the proposed convolution-like linearization strategy for pre-trained DiTs. In each text-image joint attention module, text queries aggregate information from all text and image tokens, while each image token gathers information only from tokens within a local circular window.}
   \vspace{-0.6cm}
   \label{fig:6}
\end{figure}

Unlike Neighborhood Attention and standard 2D convolution, which use a square sliding local window, CLEAR adopts circular windows, where key-value tokens within a Euclidean distance less than a predefined radius $r$ are considered for each query. 
Comparing with corresponding square windows, the computation overhead introduced by this design is $~\sim\frac{\pi}{4}$ times. 
Formally, the attention mask $M$ is constructed as follows:
\begin{equation}
    M_{ij}=\begin{cases} 
    1, & \text{if } i\leq n_{text} \text{ or } j\leq n_{text} \text{ or }d_{ij}^{(x)2}+d_{ij}^{(y)2} < r^2; \\
    0, & \text{otherwise},
    \end{cases}
\end{equation}
where $n_{text}$ denotes the number of text tokens. 
Fig.~\ref{fig:6} illustrates this paradigm. 

\subsection{Training and Optimization}

Although each query only has access to tokens within a local window, stacking multiple Transformer blocks enables each token to gradually capture holistic information—similar to the way convolutional neural networks operate. 
To promote functional consistency between models before and after fine-tuning, we employ a knowledge distillation objective during the fine-tuning process. 
Specifically, the conventional flow matching loss~\cite{esser2024scaling,lipman2022flow} is included:
\begin{equation}
    \mathcal{L}_{fm}=\Vert(\epsilon-z_0)-\epsilon_\theta(z_t,t,y)\Vert_2^2,
\end{equation}
where $z_0$ is denotes the feature of an image $x$ encoded with a pre-trained VAE encoder $\mathcal{E}(\cdot)$ while $z_t$ is its noisy version at the $t$-th timestep, $y$ is the text condition, and $\epsilon_\theta(\cdot)$ is the DiT backbone for denoising with parameters $\theta$. 
Beyond that, we encourage consistency between the linearized student model and the original teacher model, in terms of predictions and attention outputs:
\begin{equation}
\begin{aligned}
    \mathcal{L}_{pred}&=\Vert\epsilon_{\theta}(z_t,t,y)-\epsilon_{\theta_{org}}(z_t,t,y)\Vert_2^2,\\
    \mathcal{L}_{attn}=&\frac{1}{L}\sum_{l=1}^{L}\Vert\epsilon_{\theta}^{(l)}(z_t,t,y)-\epsilon_{\theta_{org}}^{(l)}(z_t,t,y)\Vert_2^2,\label{eq:5}
\end{aligned}
\end{equation}
where $\theta_{org}$ denotes parameters of the original teacher DiT, $L$ is the number of attention layers applying the loss term, and the superscript $^{(l)}$ indicates the layer index. 
The training objectives can be written as:
\begin{equation}
    \min_{\theta}\mathbb{E}_{z\sim\mathcal{E}(x),y,\epsilon\sim\mathcal{N}(0,1),t}[\mathcal{L}_{fm}+\alpha\mathcal{L}_{pred}+\beta\mathcal{L}_{attn}],\label{eq:6}
\end{equation}
where $\alpha$ and $\beta$ are hyper-parameters controlling the weights of the corresponding loss terms. 
Only parameters in the attention layers are trainable. 

For the training data, we find that training on samples generated by the original DiT model yields significantly better results than training on a real image dataset, even when the real dataset contains much more higher-quality data. 
Please refer to Sec.~\ref{sec:4-3} for more discussions. 

\begin{figure}[t]
  \centering
   \includegraphics[width=\linewidth]{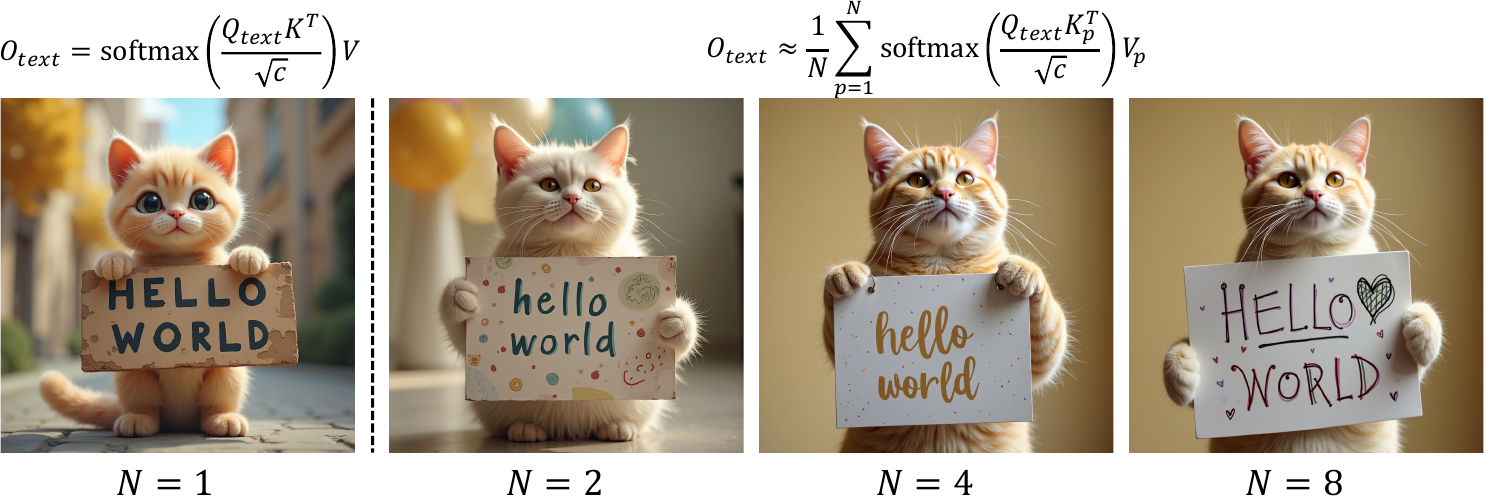}
   \vspace{-0.7cm}
   \caption{To enhance multi-GPU parallel inference, each text query aggregates only the key-value tokens from the patch managed by its assigned GPU, then averages the attention results across all GPUs, which also generates high-quality images.}
   \vspace{-0.6cm}
   \label{fig:7}
\end{figure}

\begin{table*}[!t]
    \centering
    \resizebox{\linewidth}{!}{
    \begin{tabular}{c||cccc||cc||ccc}
    \toprule
    \toprule
\multirow{2}{*}{\textbf{Method/Setting}} & \multicolumn{4}{c}{\textbf{Against Original}}                                                & \multicolumn{2}{c}{\textbf{Against Real}}            & \multirow{2}{*}{\textbf{CLIP-T} ($\uparrow$)} & \multirow{2}{*}{\textbf{IS} ($\uparrow$)} & \multirow{2}{*}{\textbf{GFLOPS} ($\downarrow$)} \\
                                & \textbf{FID} ($\downarrow$) & \textbf{LPIPS} ($\downarrow$) & \textbf{CLIP-I} ($\uparrow$) & \textbf{DINO} ($\uparrow$) & \textbf{FID} ($\downarrow$) & \textbf{LPIPS} ($\downarrow$) &                                      &                                  &                                        \\
\midrule
\midrule
Original FLUX-1.dev                       & -                  & -                    & -                   & -                 & 34.93              & 0.81                 & 31.06                                & 38.25                            & 260.9                                  \\
\midrule
\midrule
Sigmoid Attention~\cite{ramapuram2024theory}                       & 447.80              & 0.91                 & 41.34               & 0.25              & 457.69             & 0.84                 & 17.53                                & 1.15                             & 260.9                                  \\
Linear Attention~\cite{dao2024transformers,liu2024linfusion,yang2023gated,katharopoulos2020transformers}                       & 324.54             & 0.85                 & 51.37               & 2.17              & 325.58             & 0.87                 & 19.16                                & 2.91                             & 174.0                                  \\
PixArt-Simga~\cite{chen2024pixartsigma}                    & 30.64              & 0.56                 & 86.43               & 71.45             & 33.38              & 0.88                 & 31.12                                & 32.14                            & 67.7                                   \\
Agent Attention~\cite{han2025agent}                 & 69.85              & 0.65                 & 78.18               & 56.09             & 54.31              & 0.87                 & 30.38                                & 21.03                            & 80.5                                   \\
Strided Attention~\cite{child2019generating}               & 24.88              & 0.61                 & 85.50               & 70.72             & 35.27              & 0.89                 & 30.62                                & 32.05                            & 67.7                                   \\
Swin Transformer~\cite{liu2021swin}                & 18.90              & 0.65                 & 85.72               & 73.43             & 32.20              & 0.87                 & 30.64                                & 34.68                            & 67.7                                   \\
\midrule
\midrule
CLEAR ($r=8$)                          & 15.53              & 0.64                 & 86.47               & 74.36             & 32.06              & 0.83                 & 30.69                                & 34.47                            & 63.5                                   \\
\rowcolor{cvprblue!15}w. distill                      & 13.07              & 0.62                 & 88.56               & 77.66             & 33.06              & 0.82                 & 30.82                                & 35.92                            & 63.5          \\
\midrule
CLEAR ($r=16$)                          & 14.27              & 0.60                  & 88.51               & 78.35             & 32.36              & 0.89                 & 30.90                                & 37.13                            & 80.6                                   \\
\rowcolor{cvprblue!15}w. distill                      & 13.72              & 0.58                 & 88.53               & 77.30             & 33.63              & 0.88                 & 30.65                                & 37.84                            & 80.6                                   \\
\midrule
CLEAR ($r=32$)                          & 11.07              & 0.52                 & 89.92               & 81.20             & 33.47              & 0.82                 & 30.96                                & 37.80                            & 154.1                                  \\
\rowcolor{cvprblue!15}w. distill                      & 8.85               & 0.46                 & 92.18               & 85.44             & 34.88              & 0.81                 & 31.00                                & 39.12                            & 154.1                                  \\
\bottomrule
\bottomrule
\end{tabular}
    }
    \vspace{-0.3cm}
    \caption{Quantitative results of the original FLUX-1.dev, previous efficient attention methods, and CLEAR proposed in this paper with various $r$ on 5,000 images from the COCO2014 validation dataset at a resolution of $1024\times1024$.}
    \vspace{-0.7cm}
    \label{tab:main}
\end{table*}

\subsection{Multi-GPU Parallel Inference}

Since attention is confined to a local window around each query, CLEAR offers greater efficiency for multi-GPU patch-wise parallel inference compared to the full attention in the original DiTs, which is particularly valuable for generating ultra-high-resolution images. 
Specifically, each GPU is responsible for processing an image patch, and the GPU communication is only required in the boundary areas. 
In other words, if we divide a $H\times W$ feature map into $N$ patches along the vertical dimension, with each GPU handling a $\frac{H}{N}\times W$ patch, the communication cost for image tokens between each adjacent GPUs is $\mathcal{O}(r\times W)$ in CLEAR comparing with $\mathcal{O}(H\times W)$ in the original DiT. 

Nevertheless, since each text token requires information from all image tokens, the exact attention computation in CLEAR still necessitates synchronization of all key-value tokens specially for text tokens, which compromises its potential in this regard. 
Fortunately, as shown in Fig.~\ref{fig:7}, we empirically find that without any training or adaptation, the original attention computation for text tokens can be effectively approximated by a patch-wise average while not hurting the performance too much, \textit{i.e.},
\begin{equation}
    O_{text}\approx\frac{1}{N}\sum_{p=1}^{N}\mathrm{softmax}(\frac{Q_{text}K_p^\top}{\sqrt{c}})V_p,\label{eq:7}
\end{equation}
where $p$ is the patch/GPU index. 
Consequently, we only need to aggregate attention outputs for text tokens, resulting in a constant communication cost and eliminating the need to transmit all key-value tokens. 

Moreover, our pipeline is orthogonal to existing strategies for patch parallelism such as Distrifusion~\cite{li2024distrifusion}, which applies asynchronous computation and communication by using staled feature maps. 
Building CLEAR on top of these optimizations would lead to even greater acceleration. 

\section{Experiments}
\label{sec:4}

\begin{figure*}[t]
  \centering
   \includegraphics[width=\linewidth]{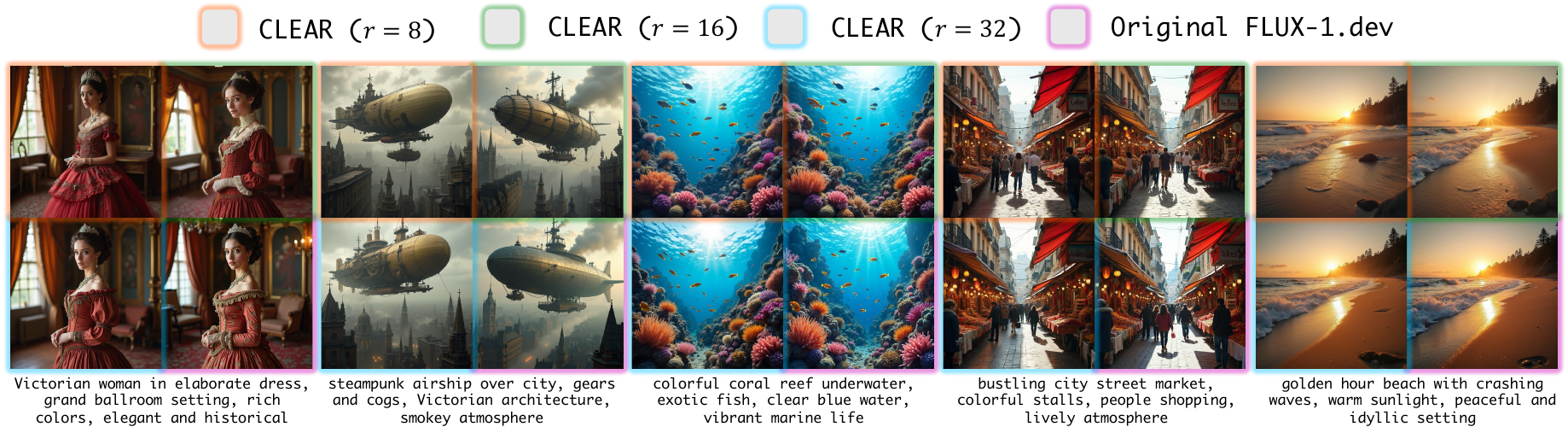}
   \vspace{-0.7cm}
   \caption{Qualitative examples by the linearized FLUX-1.dev models with CLEAR and the original model.}
   \vspace{-0.6cm}
   \label{fig:8}
\end{figure*}

\subsection{Settings and Implementation Details}

In this paper, we primarily conduct experiments with the FLUX model series due to its state-of-the-art performance in text-to-image generation. 
Studies on more DiTs can be found in the appendix. 
We replace all the attention layers in FLUX-1.dev with the proposed CLEAR and experiment with three various window sizes with $r=8$, $r=16$, and $r=32$. 
Leveraging FlexAttention in PyTorch~\cite{paszke2019pytorch}, CLEAR, as a sparse attention mechanism, can be efficiently implemented on GPUs with low-level optimizations. 
We fine-tune parameters in attention layers on 10K samples with $1024\times1024$ resolution generated by FLUX-1.dev itself\footnote{https://huggingface.co/datasets/jackyhate/text-to-image-2M/tree/main/data\_1024\_10K} for 10K iterations under a total batch size $32$ using the loss function defined in Eq.~\ref{eq:6}. 
$\mathcal{L}_{attn}$ is applied on \texttt{single\_transformer\_blocks} of FLUX, whose layer indices are $20\sim57$. 
Following previous works on architectural distillation for diffusion models~\cite{kim2023bk,liu2024linfusion}, both hyper-parameters $\alpha$ and $\beta$ are set as $0.5$. 
Other hyper-parameters, including schedulers, optimizers, \textit{etc}, follow the default settings provided by Diffusers\footnote{https://github.com/huggingface/diffusers/blob/main/examples/\\dreambooth/README\_flux.md}~\cite{von-platen-etal-2022-diffusers}. 
The training is conducted on 4 H100 GPUs supported by DeepSpeed ZeRO-2~\cite{rajbhandari2020zero}, which takes $\sim1$ day to finish. 
Unless otherwise specified, all inference is conducted on a single H100 GPU. 

Following previous works~\cite{li2024distrifusion,liu2024linfusion}, we quantitatively study the proposed method on the validation set of COCO2014~\cite{lin2014microsoft} and randomly sample 5,000 images along with their prompts for evaluation. 
Since CLEAR aims to linearize a pre-trained DiT, we also benchmark our method against the results by the original DiT using consistent random seeds. 
Following conventions~\cite{ruiz2023dreambooth,kumari2023multi,wei2023elite,ye2023ip-adapter}, we consider FID~\cite{heusel2017gans}, LPIPS~\cite{zhang2018unreasonable}, CLIP image similarity~\cite{radford2021learning}, and DINO image similarity~\cite{caron2021emerging} in this setting as metrics. 
For settings requiring pixel-wise alignment like image upsampling and ControlNet~\cite{zhang2023adding}, we additionally incorporate PSNR~\cite{hore2010image} and multi-scale SSIM~\cite{wang2003multiscale} for reference. 
While comparing with real images in COCO, we only include FID and LPIPS for distributional distances. 
Furthermore, CLIP text similarity~\cite{radford2021learning}, Inception Score (IS)~\cite{salimans2016improved}, and the number of floating point operations (FLOPS) are adopted to reflect textual alignment, general image quality, and computational burden, respectively. 
Text prompts for qualitative examples are generated by GPT-4o\footnote{https://openai.com/index/hello-gpt-4o/}. 

\subsection{Main Comparisons}

We aim to linearize a pre-trained DiT in this paper and the linearized model is expected to perform comparably with the original one. 
As illustrated in Sec.~\ref{sec:3-1}, most efficient attention algorithms result in suboptimal performance for the target problem, as quantitatively supported by the evaluation in Tab.~\ref{tab:main}. 
In contrast, the proposed convolution-like linearization strategy achieves comparable or even superior performance to the original FLUX-1.dev while requiring less computation, underscoring its potential for effectively linearizing pre-trained DiTs. 
Please refer to the appendix for an analysis of training dynamics and convergence. 

With the knowledge distillation loss terms defined in Eq.~\ref{eq:5}, the differences between outputs from the linearized models and the original model are further minimized. 
For instance, the CLIP Image score exceeds $90$ when $r=32$. 
Qualitatively, as shown in Fig.~\ref{fig:8}, the linearized models by the proposed CLEAR preserve the original outputs' overall layout, texture, and tone.

\begin{table}[!t]
    \centering
    \setlength{\tabcolsep}{2pt}
    \resizebox{\linewidth}{!}{
    \begin{tabular}{cccccccccc}
    \toprule
    \toprule
\textbf{Setting}                                 & \textbf{PSNR ($\uparrow$)} & \textbf{SSIM ($\uparrow$)} & \textbf{FID ($\downarrow$)} & \textbf{LPIPS ($\downarrow$)} & \textbf{CLIP-I ($\uparrow$)} & \textbf{DINO ($\uparrow$)} & \textbf{CLIP-T ($\uparrow$)} & \textbf{IS ($\uparrow$)} & \textbf{GFLOPS ($\downarrow$)} \\
\midrule
\midrule
\multicolumn{10}{c}{\textbf{--$\mathbf{1024\times1024\rightarrow2048\times2048}$--}}                                                                                                                                                                                                                                                                              \\
\midrule
FLUX-1.dev                              & -                          & -                          & -                           & -                             & -                            & -                          & 31.11                        & 24.53                    & 3507.9                         \\
\midrule
CLEAR ($r=8$) & 27.57                      & 0.91                       & 13.55                       & 0.12                          & 98.97                        & 98.37                      & 31.09                        & 25.05                    & 246.2                          \\
CLEAR ($r=16$)                        & 27.60                      & 0.92                       & 13.43                       & 0.12                          & 98.97                        & 98.34                      & 31.08                        & 25.46                    & 352.6                          \\
CLEAR ($r=32$)                        & 28.95                      & 0.94                       & 10.87                       & 0.10                          & 99.23                        & 98.82                      & 31.09                        & 25.48                    & 724.3                          \\
\midrule
\midrule
\multicolumn{10}{c}{\textbf{--$\mathbf{2048\times2048\rightarrow4096\times4096}$--}}                                                                                                                                                                                                                                                                              \\
\midrule
FLUX-1.dev                              & -                          & -                          & -                           & -                             & -                            & -                          & 31.29         & 24.36                    & 53604.4                        \\
\midrule
CLEAR ($r=8$) & 26.19                      & 0.87                       & 20.87                       & 0.22                         & 98.02                        & 96.56                      & 31.16                        & 25.87                    & 979.3                          \\
CLEAR ($r=16$)                        & 26.98                         & 0.88                       & 16.20                       & 0.19                         & 98.48                        & 97.64                       & 31.25                        & 25.13                    & 1433.2                         \\
CLEAR ($r=32$)                        & 27.70                      & 0.90                       & 13.56                        & 0.17                         & 98.72                        & 98.21                      & 31.20                        & 24.81                     & 3141.7                       \\
\bottomrule
\bottomrule
\end{tabular}
    }
    \vspace{-0.3cm}
    \caption{Quantitative results of the original FLUX-1.dev and our CLEAR with various $r$ on 1,000 images from the COCO2014 validation dataset at resolutions of $2048\times2048$ and $4096\times4096$.}
    \vspace{-0.7cm}
    \label{tab:hr}
\end{table}

\subsection{Empirical Studies}\label{sec:4-3}

In this part, we examine several noteworthy properties of CLEAR, including resolution extrapolation, zero-shot generalization across different models and plugins, multi-GPU parallel inference, and the effects of various training data. 

\textbf{Resolution Extrapolation.} 
One of the key advantages of a linearized diffusion model is its ability to efficiently generate ultra-high-resolution images~\cite{liu2024linfusion}. 
However, many previous studies have revealed that it is challenging for diffusion models to generate images beyond their native resolution during training.~\cite{du2024demofusion,he2023scalecrafter,bar2023multidiffusion,wu2024megafusion,du2024max}. 
They thus apply a practical solution for generating high-resolution images in a coarse-to-fine manner and devise adaptive strategies for components like position embeddings and attention scales. 
The proposed CLEAR, on the other hand, makes architectural modifications to a pre-trained diffusion backbone, making it seamlessly applicable to them. 

\begin{figure*}[t]
  \centering
   \includegraphics[width=\linewidth]{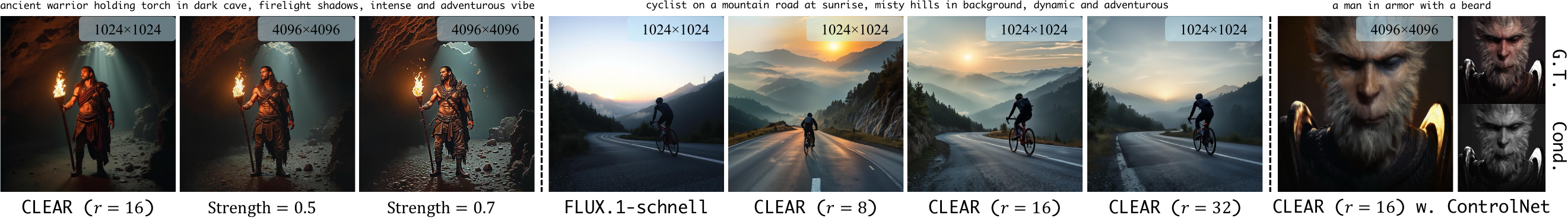}
   \vspace{-0.7cm}
   \caption{Qualitative examples of using CLEAR with SDEdit~\cite{meng2021sdedit} for high-resolution generation (left), FLUX-1.schnell in a zero-shot manner (middle), and ControlNet~\cite{zhang2023adding} (right). \texttt{G.T.} and \texttt{Cond.} denote ground-truth and condition images, separately.}
   \vspace{-0.4cm}
   \label{fig:9}
\end{figure*}

In this paper, we adopt SDEdit~\cite{meng2021sdedit}, a simple yet effective baseline adapting an image to a larger scale, for generating high-resolution images. 
In addition, we also enlarge the NTK factor of rotary position embeddings from $1$ to $10$ following \cite{peng2023ntk}, balance the entropy shift of attention using a log-scale attention factor~\cite{jin2023training}, and disable the resolution-aware dynamic shifting~\cite{esser2024scaling} in the denoising scheduler. 
By adjusting the editing strength in SDEdit, as shown in Fig.~\ref{fig:9}(left), we can effectively control the trade-off between fine details and content preservation. 
In the appendix, we also try building CLEAR on top of various methods for resolution extrapolation. 

Quantitatively, we measure the dependency between results by CLEAR with those by the original FLUX-1.dev. 
As shown in Tab.~\ref{tab:hr}, we achieve MS-SSIM scores as high as 0.9, showcasing the effectiveness of the linearized model with CLEAR as an efficient alternative to the original FLUX.

\begin{table}[!t]
    \centering
    \setlength{\tabcolsep}{2pt}
    \resizebox{\linewidth}{!}{
    \begin{tabular}{c||cccc||cc||cc}
    \toprule
    \toprule
\multirow{2}{*}{\textbf{Setting}} & \multicolumn{4}{c}{\textbf{Against Original}}                                                                           & \multicolumn{2}{c}{\textbf{Against Real}}                     & \multirow{2}{*}{\textbf{CLIP-T ($\uparrow$)}} & \multirow{2}{*}{\textbf{IS ($\uparrow$)}} \\
                                  & \textbf{FID ($\downarrow$)} & \textbf{LPIPS ($\downarrow$)} & \textbf{CLIP-I ($\uparrow$)} & \textbf{DINO ($\uparrow$)} & \textbf{FID ($\downarrow$)} & \textbf{LPIPS ($\downarrow$)} &                                               &                                           \\
                                  \midrule
FLUX-1.dev                        & -                           & -                             & -                            & -                          & 29.19                       & 0.83                          & 31.53                                         & 36.41                                     \\
\midrule
CLEAR ($r=8$)                     & 13.62                       & 0.62                          & 88.91                        & 78.36                      & 33.51                       & 0.81                          & 31.35                                         & 38.42                                     \\
CLEAR ($r=16$)                    & 12.51                       & 0.58                          & 90.43                        & 81.32                      & 34.43                       & 0.82                          & 31.38                                         & 39.66                                     \\
CLEAR ($r=32$)                    & 12.43                       & 0.57                          & 90.70                        & 82.61                      & 33.57                       & 0.83                          & 31.48                                         & 39.68                                    \\
\bottomrule
\bottomrule
\end{tabular}
    }
    \vspace{-0.3cm}
    \caption{Quantitative zero-shot generalization results to FLUX-1.schnell using CLEAR layers trained on FLUX-1.dev.}
    \vspace{-0.6cm}
    \label{tab:schnell}
\end{table}

\textbf{Cross-DiT Generalization.} 
We empirically find that the trained CLEAR layers for one DiT are also applicable for others within the same series without any further adaptation efforts. 
For example, as shown in Fig.~\ref{fig:9}(middle), the CLEAR layers trained on FLUX-1.dev can be directly applied to inference on FLUX-1.schnell, a timestep-distilled DiT supporting 4-step inference, yielding results similar to those of the original FLUX-1.schnell. 
Such zero-shot generalization is quantitatively evaluated in Tab.~\ref{tab:schnell}. 

\begin{table*}[!t]
    \centering
    \resizebox{\linewidth}{!}{
    \begin{tabular}{c||cccccc||cc||ccc}
    \toprule
    \toprule
\multirow{2}{*}{Setting} & \multicolumn{6}{c}{Against Original}                                                                                                                                              & \multicolumn{2}{c}{Against Real}                              & \multirow{2}{*}{\textbf{CLIP-T ($\uparrow$)}} & \multirow{2}{*}{\textbf{IS ($\uparrow$)}} & \multirow{2}{*}{\textbf{RMSE ($\downarrow$)}} \\
                         & \textbf{PSNR ($\uparrow$)} & \textbf{SSIM ($\uparrow$)} & \textbf{FID ($\downarrow$)} & \textbf{LPIPS ($\downarrow$)} & \textbf{CLIP-I ($\uparrow$)} & \textbf{DINO ($\uparrow$)} & \textbf{FID ($\downarrow$)} & \textbf{LPIPS ($\downarrow$)} &                                               &                                           &                                               \\
                         \midrule
                         \midrule
FLUX-1.dev               & -                          & -                          & -                           & -                             & -                            & -                          & 40.25                       & 0.32                          & 30.16                                         & 22.22                                     & 0.0385                                        \\
\midrule
CLEAR ($r=8$)            & 25.95                      & 0.93                       & 26.14                       & 0.19                          & 93.39                        & 94.24                      & 43.82                       & 0.31                          & 29.90                                         & 21.29                                     & 0.0357                                        \\
CLEAR ($r=16$)           & 28.24                      & 0.95                       & 16.86                       & 0.13                          & 96.00                        & 96.73                      & 40.45                       & 0.31                          & 30.19                                         & 22.34                                     & 0.0395                                        \\
CLEAR ($r=32$)           & 30.59                      & 0.97                       & 11.57                       & 0.09                          & 97.33                        & 98.12                      & 40.21                       & 0.31                          & 30.21                                         & 21.94                                     & 0.0419                                       \\
    \bottomrule
    \bottomrule
\end{tabular}
    }
    \vspace{-0.3cm}
    \caption{Quantitative zero-shot generalization results of the proposed CLEAR to a pre-trained ControlNet with grayscale image conditions on 1,000 images from the COCO2014 validation dataset. \texttt{RMSE} here denotes Root Mean Squared Error computed against condition images.}
    \vspace{-0.6cm}
    \label{tab:ctr}
\end{table*}

\textbf{Compatibility with DiT Plugins.} 
It is favorable that substituting original attention layers with the linearized ones would not impact the functionality of plugins trained for the original DiT. 
As shown in Fig.~\ref{fig:9}(right), CLEAR demonstrates this property by supporting the pre-trained ControlNet~\cite{zhang2023adding}, using grayscale images as conditional inputs to FLUX-1.dev\footnote{https://huggingface.co/Shakker-Labs/FLUX.1-dev-ControlNet-Union-Pro}. 
Quantitative results in Tab.~\ref{tab:ctr} further demonstrate performance comparable to the original model. 
More evaluations can be found in the appendix. 

\begin{table}[!t]
    \centering
    \setlength{\tabcolsep}{2pt}
    \resizebox{\linewidth}{!}{
    \begin{tabular}{c||cccc||cc||cc}
    \toprule
    \toprule
\multirow{2}{*}{\textbf{Setting}} & \multicolumn{4}{c}{\textbf{Against Original}}                                                                           & \multicolumn{2}{c}{\textbf{Against Real}}                     & \multirow{2}{*}{\textbf{CLIP-T ($\uparrow$)}} & \multirow{2}{*}{\textbf{IS ($\uparrow$)}} \\
                                  & \textbf{FID ($\downarrow$)} & \textbf{LPIPS ($\downarrow$)} & \textbf{CLIP-I ($\uparrow$)} & \textbf{DINO ($\uparrow$)} & \textbf{FID ($\downarrow$)} & \textbf{LPIPS ($\downarrow$)} &                                               &                                           \\
                                  \midrule
CLEAR ($r=16$)                        & -                           & -                             & -                            & -                          & 33.63                       & 0.88                          & 30.65                                         & 37.84                                     \\
\midrule
$N=2$                     & 11.55                       & 0.51                          & 90.46                        & 80.89                      & 33.74                       & 0.81                          & 31.21                                         & 39.26                                     \\
$N=4$                    & 12.78                       & 0.54                          & 89.74                        & 79.99                      & 33.07                       & 0.81                          & 31.27                                         & 40.01                                     \\
$N=8$                    & 14.21                       & 0.57                          & 88.92                        & 78.65                      & 32.26                       & 0.80                          & 31.22                                         & 39.34                                    \\
\bottomrule
\bottomrule
\end{tabular}
    }
    \vspace{-0.3cm}
    \caption{Results of patch-wise multi-GPU parallel inference with various numbers of patches using the approximation in Eq.~\ref{eq:7}.}
    \vspace{-0.7cm}
    \label{tab:patch}
\end{table}

\textbf{Multi-GPU Parallel Inference.} 
A linear complexity DiT can inherently support patch-wise multi-GPU parallel inference. 
In practice, for text-image joint attention used in modern architectures of DiTs~\cite{esser2024scaling,blackforestlabs_flux}, we have to figure out a communication-efficient solution to handle text tokens, which require gathering information from all key-value tokens. 
In Eq.~\ref{eq:7}, we propose an approximation via patch-wise average and validate the effectiveness quantitatively in Tab.~\ref{tab:patch}. 
Results indicate a high correlation in semantics before and after this approximation, demonstrating its practical effectiveness. 
Details on acceleration performance and compatibility with asynchronous communication methods like Distrifusion~\cite{li2024distrifusion} are discussed in the appendix.

\textbf{Training Data.} 
In our initial exploration, we fine-tune the FLUX model using a 200K subset of LAION5B~\cite{schuhmann2022laion}, consisting of high-quality real text-image pairs with aesthetic scores greater than 7. 
However, it would led to inferior results compared to the 10K synthetic samples generated by FLUX-1.dev itself. 
We then investigate the impact of different training data by fine-tuning the model without altering its structure. 
We find that the training dynamics differ significantly: training on synthetic data results in a much lower loss. 
We speculate that the discrepancy arises from the mismatch between the LAION dataset distribution and the data distribution for training FLUX. 
Fine-tuning on LAION data may cause the model to struggle with this distribution shift, leading to increased training difficulty.

\section{Conclusions}
\label{sec:5}

\begin{figure}[t]
  \centering
   \includegraphics[width=\linewidth]{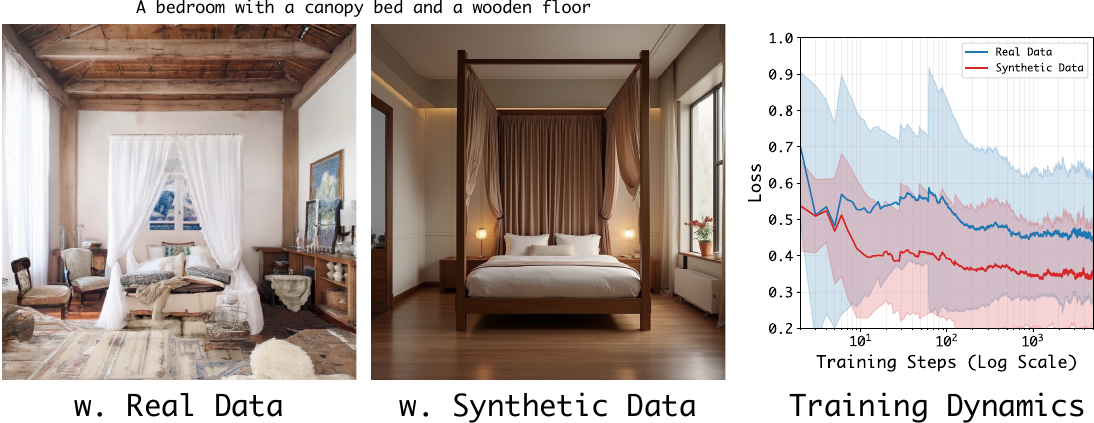}
   \vspace{-0.7cm}
   \caption{Fine-tuning on real data results in inferior performance compared to fine-tuning on self-generated synthetic data.}
   \vspace{-0.9cm}
   \label{fig:10}
\end{figure}

In this paper, we present CLEAR, a convolution-like local attention strategy that effectively linearizes the attention mechanism in pre-trained Diffusion Transformers (DiTs), making them significantly more efficient for high-resolution image generation. 
By comprehensively analyzing existing efficient attention methods, we identified four key elements—locality, formulation consistency, high-rank attention maps, and feature integrity—that are essential for successful linearization in the context of pre-trained DiTs. 
CLEAR leverages these principles by restricting attention to a circular local window around each query, achieving linear complexity while retaining high-quality results comparable to the original model. 
Our experiments demonstrate that fine-tuning on merely 10K self-generated samples allows for efficient knowledge transfer to a student model, leading to a 99.5\% reduction in attention computations and a $6.3\times$ acceleration in 8K-resolution image generation. 
Moreover, CLEAR’s distilled attention layers support zero-shot generalization across different models and plugins and improve multi-GPU parallel inference capabilities, offering broader applicability and scalability. 

One limitation of our approach is that, the practical acceleration achieved by CLEAR does not fully meet the theoretical expectations indicated by FLOPS. 
It becomes less significant at relatively low resolutions and can even be slower than the original DiT when the resolution is below $1024\times1024$. 
This drawback arises partially because hardware optimization for sparse attention is inherently more challenging than the optimizations achieved by FlashAttention for full attention computation. 
Addressing this limitation may require developing fused CUDA operators specifically optimized for the specific sparse pattern of CLEAR, which is a valuable direction for future works. 

\section*{Acknowledgments}
We would like to acknowledge that the computational work involved in this research work is partially supported by NUS IT’s Research Computing group using grant numbers NUSREC-HPC-00001. 

{
    \small
    \bibliographystyle{ieeenat_fullname}
    \bibliography{main}
}

\clearpage

\appendix

\section{Details of Efficient Attention Alternatives}

The vanilla scaled dot-product attention, although effective and flexible, introduces quadratic computational complexity. 
Many works have focused on its efficient alternatives. 
In Sec.~\ref{sec:2}, we provide a taxonomic overview of recent works and will supplement more details regarding the specific formulations and implementations here. 

\textbf{Linear Attention}  
avoids the \texttt{softmax} operation in the vanilla attention, supporting computing $K^\top V$ first with the associative property of matrix-wise multiplication, and thus achieves linear complexity. 
Before that, non-negative kernel functions $f(\cdot)$ and $g(\cdot)$ are applied on $Q$ and $K$ respectively such that the similarity between each query-key pair is non-negative. 
Furthermore, the similarity score between each query-key pair is normalized by the sum of similarity scores of between this query and all key tokens separately, to mimic the functionalities of \texttt{softmax}. 
Following \cite{katharopoulos2020transformers,han2024demystify,liu2024linfusion}, we implement $f(\cdot)$ and $g(\cdot)$ by the $\mathrm{elu}$ function~\cite{clevert2015fast}. 
Formally, the operation for the $i$-th query can be written as\footnote{https://github.com/LeapLabTHU/MLLA}:
\begin{equation}
    O_i=\frac{(\mathrm{elu}(Q_i)+1)(\mathrm{elu}(K)+1)^\top}{(\mathrm{elu}(Q_i)+1)\sum_{j=1}^m(\mathrm{elu}(K_j)+1)^\top}V.
\end{equation}

\textbf{Sigmoid Attention} 
replaces the \texttt{softmax} with the formulation of \texttt{sigmoid}, which removes the need to compute the \texttt{softmax} normalization, and thus achieves acceleration:
\begin{equation}
    O=\mathrm{sigmoid}(\frac{QK^\top}{\sqrt{c}}+b)V,
\end{equation}
where $b$ is a hyper-parameter. 
In this paper, we follow the official implementation of \texttt{FlashSigmoid} with hardware-aware optimization\footnote{https://github.com/apple/ml-sigmoid-attention} when applying Sigmoid Attention to DiTs. 

\textbf{PixArt-Sigma} 
achieves acceleration by spatially down-sampling the key-value token maps~\cite{chen2024pixartsigma}. 
Following the official implementation\footnote{https://github.com/PixArt-alpha/PixArt-sigma}, we use learnable group-wise $\mathrm{Conv}4\times4$ kernels with $\mathrm{stride}=4$ and initialize the weights to $\frac{1}{16}$ so that it is equivalent to an average pooling operation at the beginning. 
Formally, it can be written as:
\begin{equation}
    O=\mathrm{softmax}(\frac{Q\mathrm{Conv}_k(K)^\top}{\sqrt{c}})\mathrm{Conv}_v(V). 
\end{equation}
Although it has been demonstrated that such a strategy can work well at relatively deep layers of DiTs, the results are still unsatisfactory for a completely linearized DiT. 

\textbf{Agent Attention} 
performs efficient attention operations via agent tokens $A$ from a down-sampled query token map~\cite{han2025agent}:
\begin{equation}
    A=\mathrm{softmax}(\frac{\mathrm{Down}(Q)K^\top}{\sqrt{c}})V. 
\end{equation}
The derived agent tokens $A$ are then used as value tokens:
\begin{equation}
    O=\mathrm{softmax}(\frac{Q\mathrm{Down}(Q)^\top}{\sqrt{c}})A.
\end{equation}
Such operations can be viewed as an adaptive token down-sampling strategy. 

\textbf{Slot Attention} 
implemented in this paper is adapted from \cite{zhang2024gated,peng2021abc}, which contain $s$ key-value memory slots derived by adaptively aggregating key-value tokens:
\begin{equation}
    \tilde{K}=\mathrm{softmax}(\frac{PX^\top}{\sqrt{c}})K,\quad \tilde{V}=\mathrm{softmax}(\frac{PX^\top}{\sqrt{c}})V, 
\end{equation}
where $P\in\mathbb{R}^{s\times c}$ is learnable and introduced for modeling the writing intensity of each input token to each memory slot. 
These slots are then used as alternatives to original key-value tokens for attention computation:
\begin{equation}
    O=\mathrm{softmax}(\frac{Q\tilde{K}^\top}{\sqrt{c}})\tilde{V}.
\end{equation}
This strategy presents a different fashion for adaptive key-value compression. 

\textbf{Strided Attention} 
samples tokens at a regular interval~\cite{child2019generating}. 
As a sparse attention strategy, the attention mask of the $l$-th layer with a down-sampling ratio of $r\times r$ can be constructed in the following way:
\begin{equation}
    M^{(l)}_{ij}=\begin{cases} 
    1, & \text{if } i\leq n_{text} \text{ or } j\leq n_{text} \text{ or }\\ & (d_{ij}^{(x)}\%r=r_x \text{ and }d_{ij}^{(y)}\%r=r_y); \\
    0, & \text{otherwise},
    \end{cases}
\end{equation}
where $r_x=l\%r$ and $r_y=l//r$ ensure that each token has chances to be sampled as key-value tokens. 

\textbf{Swin Transformer} 
adopts a sliding window partition strategy~\cite{liu2021swin}, where attention interactions are independently conducted for each window. 
Formally, the attention map can be constructed via:
\begin{equation}
    M_{ij}=\begin{cases} 
    1, & \text{if } i\leq n_{text} \text{ or } j\leq n_{text} \text{ or }\\ &\text{ tokens }i\text{ and }j\text{ in the same window}; \\
    0, & \text{otherwise}.
    \end{cases}
\end{equation}
We set the window size to $16$ and apply a shift of $8$ for windows in layers with odd indices, following the approach described in the original manuscript. 
The rank of the image-to-image attention mask corresponds to the number of windows, which poses challenges in achieving the high-rank requirement introduced in the main manuscript needed for linearizing DiTs. 

\begin{figure}[t]
  \centering
   \includegraphics[width=\linewidth]{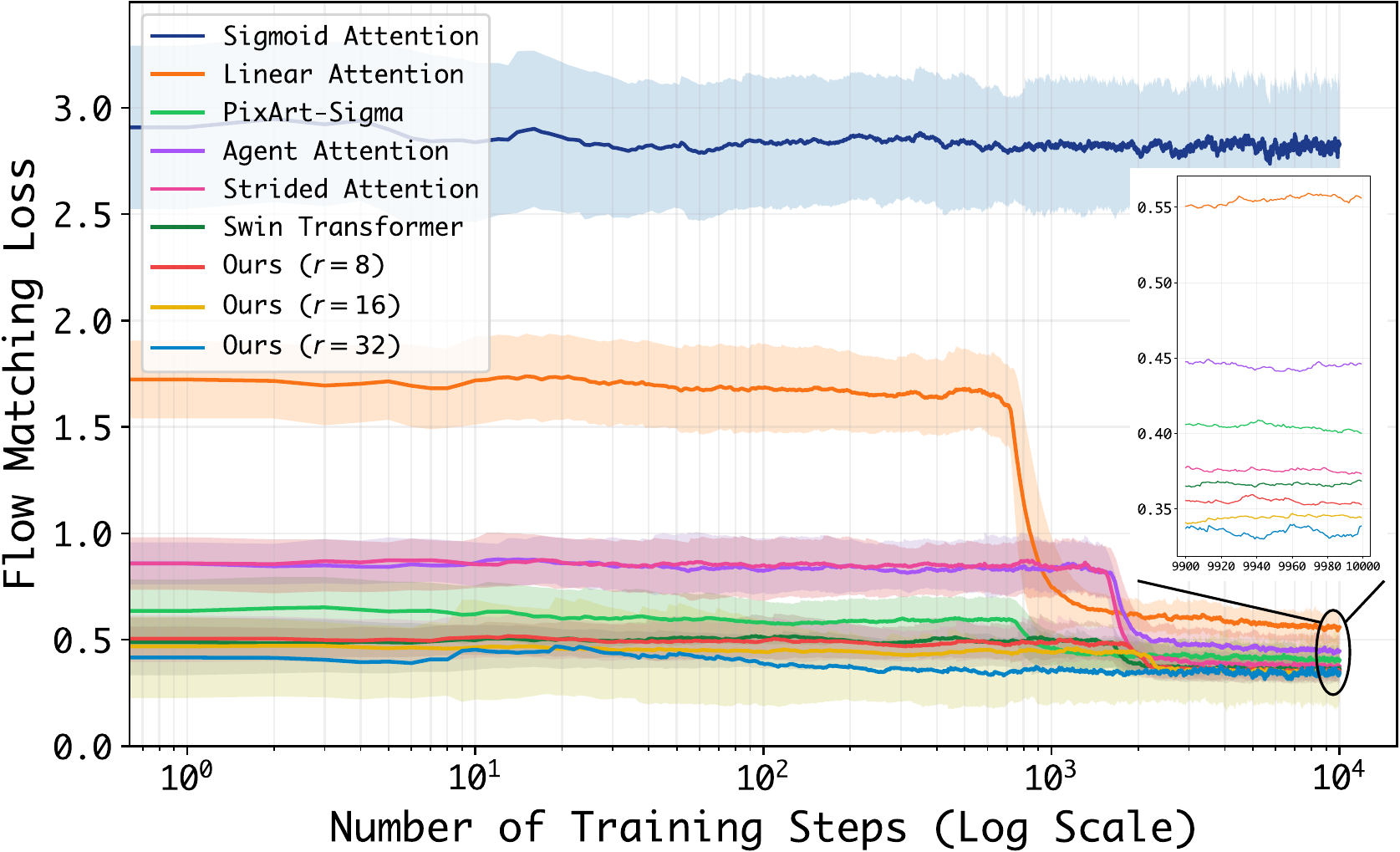}
   \vspace{-0.7cm}
   \caption{Training dynamics of various efficient attention alternatives on FLUX-1.dev.}
   \vspace{-0.3cm}
   \label{fig:11}
\end{figure}

\section{Training Dynamics}

We supplement the curves of training losses of various efficient attention alternatives in Fig.~\ref{fig:11}. 
The conclusion is consistent with the main manuscript, that strategies fulfilling the requirements of locality, formulation consistency, high-rank attention map, and feature integrity yield the most satisfactory training convergence.  

\section{Raw Data for Efficiency Comparisons}

We supplement raw data for Fig.~\ref{fig:2} on efficiency comparisons in Tab.~\ref{tab:eff} in sake of better clarity. 

\section{Results on More DiTs}

We additionally deploy our method on DiT models other than FLUX used in the main manuscript to demonstrate the universality of the proposed CLEAR. 
Here, we consider StableDiffusion3.5-Large\footnote{https://huggingface.co/stabilityai/stable-diffusion-3.5-large}~\cite{esser2024scaling} (SD3.5-L), another state-of-the-art text-to-image generation DiT. 
We use the default setting of $r=16$, which yields the best trade-off between quality and efficiency according to our experiments. 
Results on the COCO2014 validation dataset are shown in Tab.~\ref{tab:sd3}. 

\begin{figure*}[t]
  \centering
   \includegraphics[height=0.96\textheight]{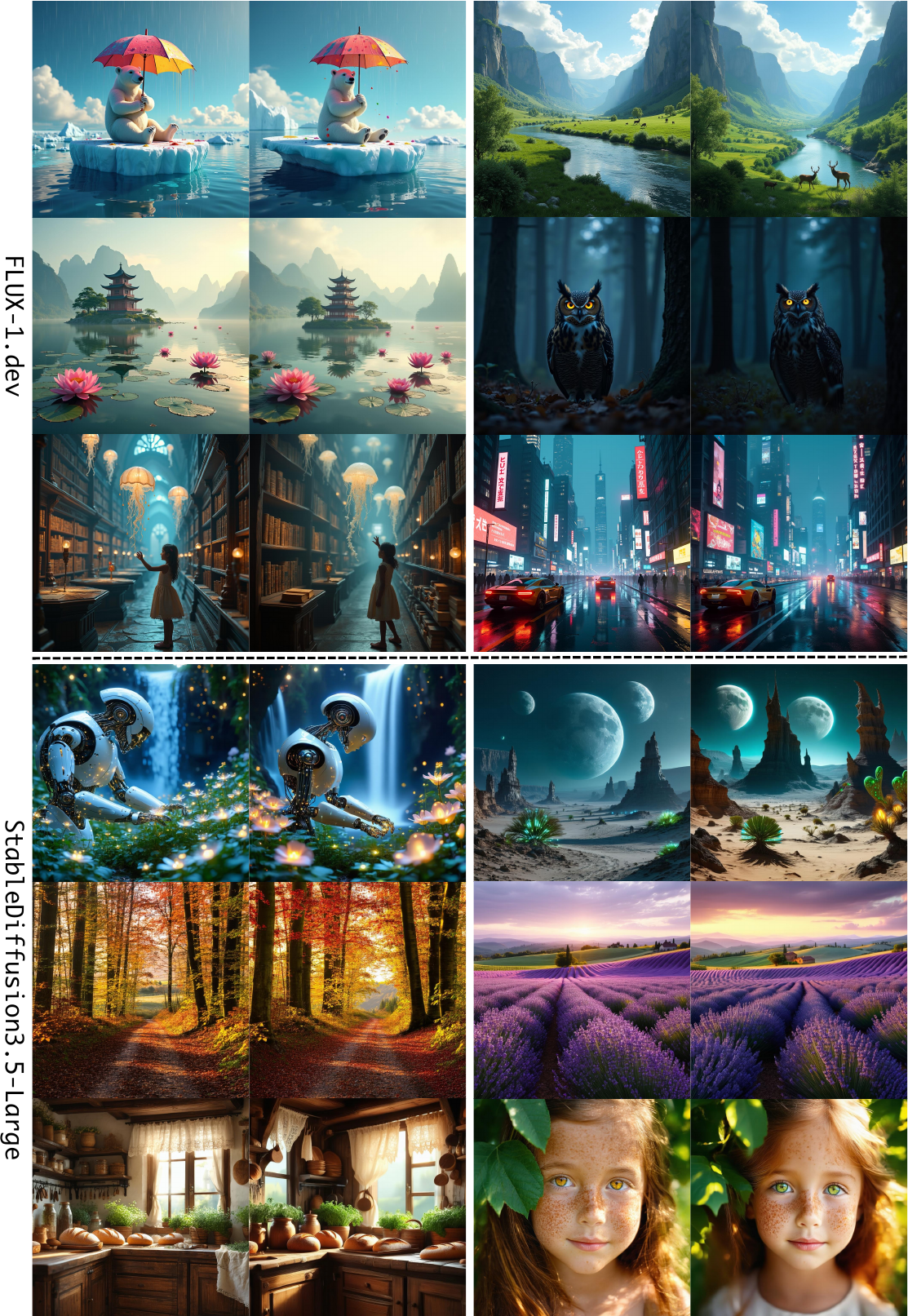}
   \vspace{-0.2cm}
   \caption{Qualitative comparisons on FLUX-1.dev (top) and SD3.5-Large (bottom). The left subplots are results by the original models while the right ones are by the CLEAR linearized models. Prompts are listed in Fig.~\ref{lst:prompt-supp}.}
   \vspace{-0.4cm}
   \label{fig:13}
\end{figure*}

We also supplement more qualitative comparisons with results by the original FLUX-1.dev and SD3.5-L in Fig.~\ref{fig:13}. 
Results indicate an overall comparable performance. 
Due to the absence of explicit long-distance token interactions, our method may underperform in capturing overall structural properties, such as potential symmetry. 
Additionally involving more or less global tokens, such as down-sampled tokens as employed in PixArt-Sigma~\cite{chen2024pixartsigma}, could potentially mitigate this issue.
However, as the primary objective of this paper is to highlight the significance of locality as a simple yet effective baseline, we leave detailed design explorations to future work. 

\begin{figure}[t]
  \centering
   \includegraphics[width=\linewidth]{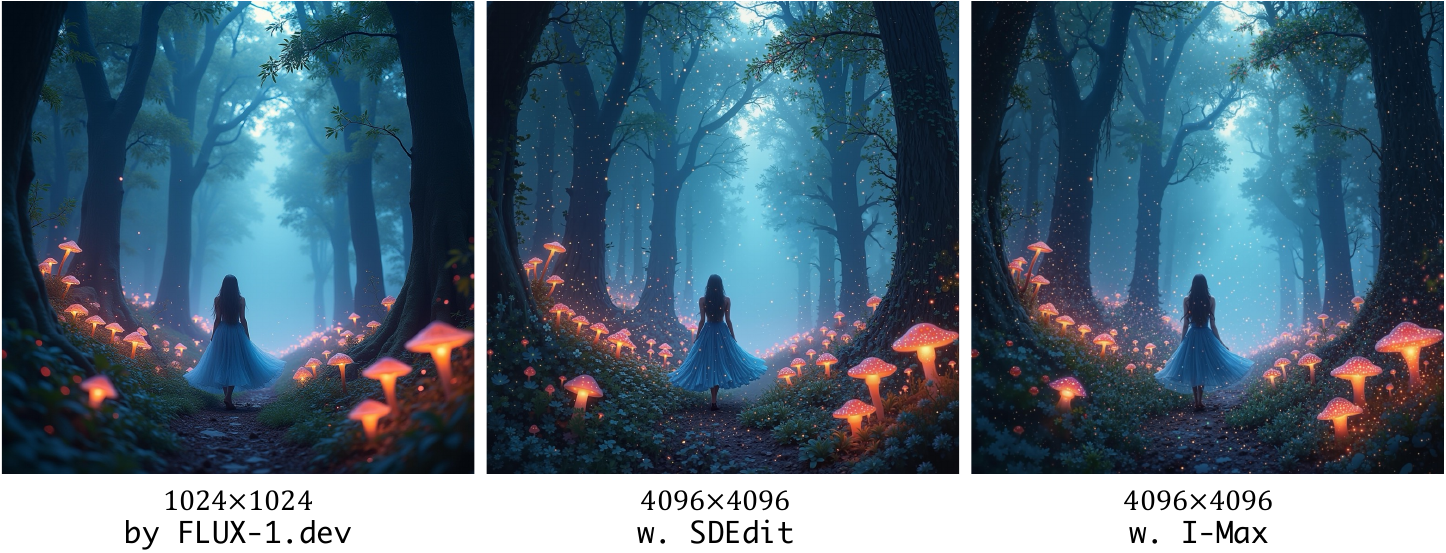}
   \vspace{-0.7cm}
   \caption{The linearized DiTs by CLEAR are compatible with various pipelines dedicated for high-resolution inference. The prompt is shown in Fig.~\ref{lst:prompt}.}
   \vspace{-0.5cm}
   \label{fig:12}
\end{figure}

\section{More High-Resolution Results}

In the main manuscript, we build our CLEAR on top of SDEdit~\cite{meng2021sdedit}, a simple yet effective strategy for image generation given a conditional image, for coarse-to-fine high-resolution generation. 
We demonstrate here that our method is also compatible with a variety of pipelines dedicated for resolution extrapolation. 
As shown in Fig.~\ref{fig:12}, we deploy CLEAR on I-Max~\cite{du2024max}, a concurrent work for training-free high-resolution generation with pre-trained DiTs, and observe that it may yield a more optimal balance between preserving low-resolution content and capturing high-resolution details. 
For instance, as shown in Fig.~\ref{fig:12}, I-Max preserves the textures of the dresses from the low-resolution result with minimal variation while effectively enhancing clear high-resolution details. 

\section{More ControlNet Results}

We conduct additional experiments with pre-trained ControlNets~\cite{zhang2023adding} to demonstrate the zero-shot generalizability of the trained CLEAR layers. 
Results for tiled image conditions and blur image conditions are shown in Tab.~\ref{tab:ctr-supp}. 

\section{Efficient Multi-GPU Parallel Inference}

For CLEAR, since there are only token interactions in the boundary areas of the patch handled by each GPU, and the approximation of feature aggregation for text tokens defined in Eq.~\ref{eq:7}, we achieve satisfactory efficiency on multi-GPU parallel inference. 
Furthermore, we can adopt the asynchronous communication strategy in Distrifusion~\cite{li2024distrifusion} to achieve even more significant acceleration. 
As shown in Tab.~\ref{tab:distri}, the acceleration become more significant with the increasing of image resolution, while the original DiT encounters out-of-memory (OOM) error due to the necessity of caching all key-value tokens. 

\begin{table*}[!t]
    \centering
    \resizebox{\linewidth}{!}{
    \begin{tabular}{c||cccc||cccc}
    \toprule
    \toprule
\multirow{2}{*}{\textbf{Setting}} & \multicolumn{4}{c}{\textbf{Running Time (Sec. / 50 Steps)}}                                                   & \multicolumn{4}{c}{\textbf{TFLOPS / Layer}}                                                                   \\
                                  & \textbf{$1024\times1024$} & \textbf{$2048\times2048$} & \textbf{$4096\times4096$} & \textbf{$8192\times8192$} & \textbf{$1024\times1024$} & \textbf{$2048\times2048$} & \textbf{$4096\times4096$} & \textbf{$8192\times8192$} \\
                                  \midrule
                                  \midrule
FLUX-1.dev                        & 4.45                      & 20.90                     & 148.97                    & 1842.48                   & 0.26                      & 3.51                      & 53.60                     & 847.73                    \\
\midrule
CLEAR ($r=8$)                     & 4.40                      & 15.67                     & 69.41                     & 293.50                    & 0.06                      & 0.25                      & 0.98                      & 3.92                      \\
CLEAR ($r=16$)                    & 4.56                      & 17.19                     & 83.13                     & 360.83                    & 0.09                      & 0.35                      & 1.43                      & 5.79                      \\
CLEAR ($r=32$)                    & 5.45                      & 19.95                     & 109.57                    & 496.22                    & 0.15                      & 0.72                      & 3.14                      & 13.09                    \\
\bottomrule
\bottomrule
\end{tabular}
    }
    \vspace{-0.3cm}
    \caption{Raw data for Fig.~\ref{fig:2} on efficiency comparisons.}
    \label{tab:eff}
\end{table*}

\begin{table*}[!t]
    \centering
    \resizebox{\linewidth}{!}{
    \begin{tabular}{c||cccc||cc||ccc}
    \toprule
    \toprule
\multirow{2}{*}{\textbf{Method/Setting}} & \multicolumn{4}{c}{\textbf{Against Original}}                                                & \multicolumn{2}{c}{\textbf{Against Real}}            & \multirow{2}{*}{\textbf{CLIP-T} ($\uparrow$)} & \multirow{2}{*}{\textbf{IS} ($\uparrow$)} & \multirow{2}{*}{\textbf{GFLOPS} ($\downarrow$)} \\
                                & \textbf{FID} ($\downarrow$) & \textbf{LPIPS} ($\downarrow$) & \textbf{CLIP-I} ($\uparrow$) & \textbf{DINO} ($\uparrow$) & \textbf{FID} ($\downarrow$) & \textbf{LPIPS} ($\downarrow$) &                                      &                                  &                                        \\
\midrule
\midrule
SD3.5-L                         & -            & -              & -               & -             & 34.10              & 0.83                 & 31.40                            & 36.06                        & 206.5                            \\
w. CLEAR ($r=16$)                 & 11.21        & 0.57           & 90.9            & 81.47         & 36.98              & 0.83                 & 31.23                            & 36.28                        & 63.8                            \\
\bottomrule
\bottomrule
\end{tabular}
    }
    \vspace{-0.3cm}
    \caption{Quantitative results of the original SD3-Large and its linearized version by CLEAR proposed in this paper on 5,000 images from the COCO2014 validation dataset at a resolution of $1024\times1024$.}
    \label{tab:sd3}
\end{table*}

\begin{table*}[!t]
    \centering
    \setlength{\tabcolsep}{2pt}
    \resizebox{\linewidth}{!}{
    \begin{tabular}{cc||cccccc||ccc||cc}
\toprule
\toprule
\multirow{2}{*}{\textbf{Condition}} & \multirow{2}{*}{\textbf{Setting}} & \multicolumn{6}{c}{\textbf{Against Original}}                                                                                                                                     & \multicolumn{2}{c}{\textbf{Against GT}}                     & \multirow{2}{*}{\textbf{CLIP-T ($\uparrow$)}} & \multirow{2}{*}{\textbf{IS ($\uparrow$)}} & \multirow{2}{*}{\textbf{RMSE ($\downarrow$)}} \\
                                    &                                   & \textbf{PSNR ($\uparrow$)} & \textbf{SSIM ($\uparrow$)} & \textbf{FID ($\downarrow$)} & \textbf{LPIPS ($\downarrow$)} & \textbf{CLIP-I ($\uparrow$)} & \textbf{DINO ($\uparrow$)} & \textbf{FID ($\downarrow$)} & \textbf{LPIPS ($\downarrow$)} &                                               &                                           &                                               \\
\midrule
\midrule
\multirow{2}{*}{Tile}               & FLUX-1.dev                        & -                          & -                          & -                           & -                             & -                            & -                          & 38.20                       & 0.31                          & 30.16                                         & 21.54                                     & 0.019                                         \\
                                    & CLEAR ($r=16$)                    & 30.12                      & 0.97                       & 9.1                         & 0.13                          & 99.25                        & 99.04                      & 39.73                       & 0.34                          & 30.11                                         & 21.77                                     & 0.021                                         \\
\midrule
\multirow{2}{*}{Blur}               & FLUX-1.dev                        & -                          & -                          & -                           & -                             & -                            & -                          & 38.72                       & 0.31                          & 30.20                                         & 21.42                                     & 0.028                                         \\
                                    & CLEAR ($r=16$)                    & 28.92                      & 0.96                       & 10.56                       & 0.13                          & 99.02                        & 98.67                      & 39.66                       & 0.33                          & 30.14                                         & 21.67                                     & 0.033                     \\

\bottomrule
\bottomrule
\end{tabular}
    }
    \vspace{-0.3cm}
    \caption{Quantitative zero-shot generalization results of the proposed CLEAR to a pre-trained ControlNet with tiled image conditions and blur image conditions on 1,000 images from the COCO2014 validation dataset. \texttt{RMSE} here denotes Root Mean Squared Error computed against condition images.}
    \label{tab:ctr-supp}
\end{table*}

\begin{table*}[!t]
    \centering
    \resizebox{\linewidth}{!}{
    \begin{tabular}{c||ccc||ccc}
    \toprule
    \toprule
\multirow{2}{*}{\textbf{\# of GPUs}} & \multicolumn{3}{c}{\textbf{Synchronous}}                               & \multicolumn{3}{c}{\textbf{Asynchronous}}                              \\
                                    & \textbf{FLUX-1.dev} & \textbf{CLEAR ($r=16$)} & \textbf{CLEAR ($r=8$)} & \textbf{FLUX-1.dev} & \textbf{CLEAR ($r=16$)} & \textbf{CLEAR ($r=8$)} \\
                                    \midrule
\midrule
\multicolumn{7}{c}{--$\mathbf{1024\times1024}$--}                                        
\\
\midrule
\midrule
1                                   & 11.13               & 11.40                   & 11.00                  & -                   & -                       & -                      \\
\midrule
2                                   & 7.98\upfast{$\times$1.39}   & 8.52\upfast{$\times$1.34}       & 7.85\upfast{$\times$1.40}      & 7.64\upfast{$\times$1.46}   & 8.10\upfast{$\times$1.41}        & 7.50\upfast{$\times$1.47}       \\
4                                   & 5.93\upfast{$\times$1.88}   & 6.01\upfast{$\times$1.90}       & 5.38\upfast{$\times$2.04}      & 5.64\upfast{$\times$1.97}   & 5.67\upfast{$\times$2.01}       & 5.11\upfast{$\times$2.15}      \\
8                                   & 4.84\upfast{$\times$2.30}   & NA                      & 4.37\upfast{$\times$2.52}      & 4.49\upfast{$\times$2.48}   & NA                      & 3.90\upfast{$\times$2.82}       \\
\midrule
\midrule
\multicolumn{7}{c}{--$\mathbf{2048\times2048}$--}                                                                                                                                     \\
\midrule
\midrule
1                                   & 52.25               & 42.98                   & 39.18                  & -                   & -                       & -                      \\
\midrule
2                                   & 30.96\upfast{$\times$1.69}  & 26.26\upfast{$\times$1.64}      & 23.96\upfast{$\times$1.64}     & 30.17\upfast{$\times$1.73}  & 25.41\upfast{$\times$1.69}      & 23.01\upfast{$\times$1.70}     \\
4                                   & 18.94\upfast{$\times$2.76}  & 15.64\upfast{$\times$2.75}      & 13.86\upfast{$\times$2.83}     & 18.58\upfast{$\times$2.81}  & 15.12\upfast{$\times$2.84}      & 13.4\upfast{$\times$2.92}      \\
8                                   & 12.97\upfast{$\times$4.03}  & 9.72\upfast{$\times$4.42}       & 8.40\upfast{$\times$4.66}       & 12.57\upfast{$\times$4.16}  & 9.30\upfast{$\times$4.62}        & 8.04\upfast{$\times$4.87}      \\
\midrule
\midrule
\multicolumn{7}{c}{--$\mathbf{4096\times4096}$--}                                                                                                                                     \\
\midrule
\midrule
1                                   & 372.43              & 207.83                  & 173.53                 & -                   & -                       & -                      \\
\midrule
2                                   & 200.16\upfast{$\times$1.86} & 115.02\upfast{$\times$1.81}     & 96.65\upfast{$\times$1.80}     & OOM                 & 112.34\upfast{$\times$1.85}     & 91.84\upfast{$\times$1.89}     \\
4                                   & 105.59\upfast{$\times$3.53} & 59.65\upfast{$\times$3.48}      & 49.70\upfast{$\times$3.49}      & OOM                 & 57.42\upfast{$\times$3.62}      & 48.57\upfast{$\times$3.57}     \\
8                                   & 59.18\upfast{$\times$6.29}  & 32.33\upfast{$\times$6.43}      & 26.88\upfast{$\times$6.46}     & OOM                 & 31.23\upfast{$\times$6.65}      & 26.26\upfast{$\times$6.61}    \\
\bottomrule
\bottomrule
\end{tabular}
    }
    \vspace{-0.3cm}
    \caption{Efficiency of multi-GPU parallel inference measured by sec./50 denoising steps on a HGX H100 8-GPU server. We adapt Distrifusion~\cite{li2024distrifusion} to FLUX-1.dev here for asynchronous communication. The ratios of acceleration are highlighted with red. Results of CLEAR with $r=16$ at the $1024\times1024$ resolution are not available (NA) because the patch size processed by each GPU is smaller than the boundary size. OOM denotes encountering out-of-memory error.}
    \label{tab:distri}
\end{table*}

\begin{figure*}[t]
  \centering
   \includegraphics[height=0.96\textheight]{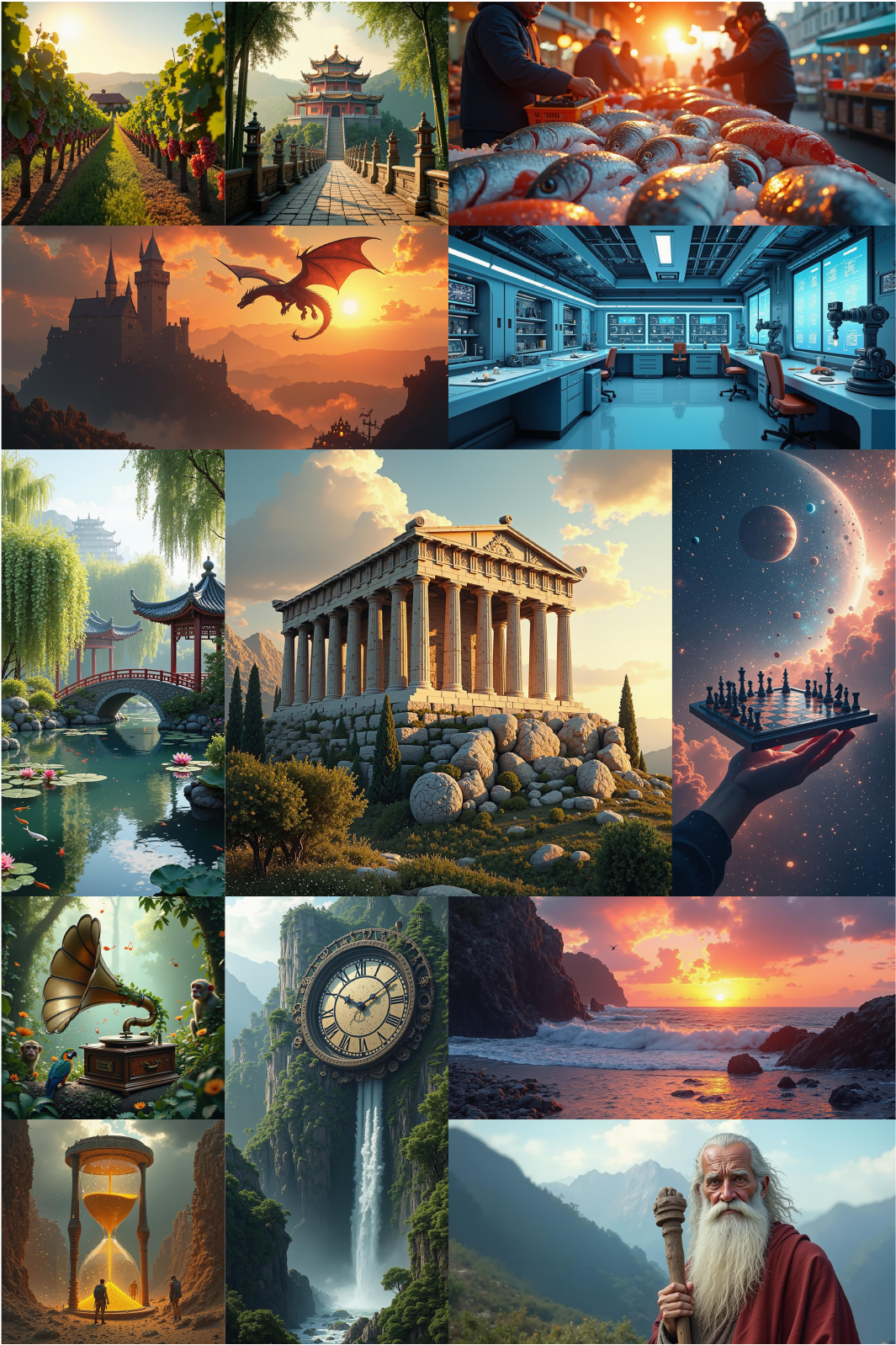}
   \vspace{-0.2cm}
   \caption{More 4K examples by the CLEAR linearized FLUX-1.dev. Prompts are listed in Fig.~\ref{lst:prompt-supp}.}
   \vspace{-0.4cm}
   \label{fig:14}
\end{figure*}

\begin{figure*}[t]
\centering
\definecolor{backgroundcolor}{RGB}{250,250,250}
\begin{minipage}{\textwidth}
\lstset{
    language=xml,
    basicstyle=\tiny\ttfamily,
    backgroundcolor=\color{backgroundcolor},
    xleftmargin=0pt,
    breaklines=true,
    frame=single,
    showstringspaces=false,
    keepspaces=true
}
\begin{lstlisting}
    // Fig. 1, according to the top-left corner, from top to bottom, from left to right

    // 1, also used in Fig. 4 and Fig. 6
    "A high fantasy scene where a fierce battle is taking place in the sky between dragons and powerful wizards. One side of the scene shows wizards casting spells, their staffs glowing with magical energy, while on the other, dragons with scales of fire and lightning breathe torrents of flame. The sky is torn with storms of magic, and below, a medieval kingdom watches in awe as the skies blaze with the fury of battle.",
    // 2
    "futuristic cityscape, towering skyscrapers, neon lights, speeding cars, holographic advertisements, cyberpunk, ultra-realistic, high resolution, cinematic lighting, highly detailed, ultra HD, 8K, nighttime, rain-soaked streets, reflections on glass, vibrant colors, misty atmosphere",
    // 3
    "A tiger is kissing a rabbit",
    // 4
    "classic fountain pen with detailed engravings, glass ink bottle with reflections, subtle ink stains, warm lighting, rich wood desk, soft shadows, high detail on pen and bottle, ultra-realistic textures, vintage and refined, calm and artistic feel, close-up, high resolution, deep blue and golden accents",
    // 5
    "beautiful Chinese woman in hanfu, surrounded by blooming peonies, flowing silk robes, elegant and ethereal, soft lighting, pastel colors, highly detailed fabric textures, delicate hair ornaments, peony petals in the air, graceful pose, traditional hairpin",
    // 6
    "charming countryside cottage, early morning sunlight, mist in the air, lush garden, rustic and cozy, ivy-covered walls, wooden fence, high detail, ultra-realistic, peaceful atmosphere, blooming flowers, warm light, quiet and serene",
    // 7
    "futuristic racing car, sleek design, neon underglow, high-speed action, dust trail, dynamic motion blur, cinematic lighting, high resolution, ultra-realistic, ultra HD, 8K, dark background, neon lights, sparks flying, intense colors, reflections on car surface",
    // 8
    "majestic Chinese dragon, swirling clouds, water and ink effect, powerful presence, dynamic and dramatic, monochromatic ink wash, swirling motion, high detail on dragon scales, whirlwind of clouds, dragon's fierce eyes, ink splashes, ancient mystical aura",
    // 9
    "traditional Chinese night market, red lanterns, crowded stalls, vibrant atmosphere, warm and lively, golden lighting, realistic and bustling, intricate market details, traditional snacks, merchants in robes, lanterns casting glow, animated crowd in background",
    // 10, also used in Fig. 2 (Appendix)
    "enchanted forest, glowing plants, towering ancient trees, a mystical girl, magical aura, fantasy style, vibrant colors, ethereal lighting, bokeh effect, ultra-detailed, painterly, ultra HD, 8K, soft glowing lights, mist and fog, otherworldly ambiance, glowing mushrooms, sparkling particles",
    // 11
    "portrait of an elderly female artist with silver hair, gentle smile, wearing glasses and colorful scarf, soft studio lighting, high detail wrinkles, ultra-realistic, warm lighting, creative and thoughtful, calm and wise, subtle background, rich textures, peaceful and inviting, close-up",
    // 12
    "futuristic soldier, robotic armor, high-tech weapon, visor with digital HUD, dark sci-fi, highly detailed, cinematic lighting, dynamic pose, ultra-realistic, ultra HD, 8K, neon accents, dark background, glowing HUD, intense expression, battle scars on armor",
    // 13
    "A watercolor-style sign reading 'Hello CLEAR' with soft gradients of blue, green, and purple, textured lettering, and subtle paint splashes",
    // 14
    "hidden paradise with peach blossoms, flowing river, distant mountains, quaint cottages, dreamlike and serene, vibrant colors, soft and warm lighting, idyllic landscape, blossoming peach trees, mist over river, villagers in traditional attire, sunlight filtering through petals",
    // 15
    "Parisian street at night, iconic street lights, cobblestone path, view of Eiffel Tower, vibrant city atmosphere, warm tones, rain reflections on street, historic architecture, romantic ambiance, ultra-realistic details, cinematic lighting, urban scene, high resolution",
    // 16
    "astronaut meeting alien creatures, cosmic background, colorful nebula, stars in background, high detail spacesuit, atmospheric lighting, sci-fi setting, calm and peaceful, otherworldly creatures, ultra-realistic, adventure in space, detailed environment",
    // 17
    "rustic wooden cabin interior, cozy and warm, fireplace glowing, wooden beams, vintage furniture, soft light from a window, warm and earthy tones, ultra-realistic details, rich textures, cozy blankets and cushions, peaceful ambiance, high resolution, natural wood grain visible",
    // 18
    "Chinese ink landscape painting, misty mountains, winding rivers, ancient pine trees, traditional ink wash painting, soft brushstrokes, monochromatic, ethereal and timeless, light mist among mountains, small thatched pavilion, subtle gradation of ink, natural flow",
    // 19
    "phoenix rising from flames, vibrant feathers, traditional Chinese mythological style, vivid and majestic, dynamic colors, dramatic lighting, intricate feather details, golden flames, radiant plumage, traditional patterns on wings, sense of rebirth",
    // 20
    "city street on a rainy day, wet pavement with reflections, people under umbrellas, soft city lights reflecting in water puddles, detailed raindrops, warm and cozy tones, misty atmosphere, ultra-realistic details, vibrant and deep colors, high contrast, peaceful rain ambiance, soft shadows, street lights glowing",
    // 21
    "ancient Chinese academy, surrounded by bamboo forest, stone paths, wooden study desks, calm and serene, warm lighting, natural greens, intricate woodwork, rustic textures, bamboo shadows on ground, calligraphy brushes, traditional scrolls, scholars in robes",
    // 22
    "1950s American diner, red leather booths, checkerboard floor, neon signs, nostalgic atmosphere, warm lighting, retro decor, vintage menu, chrome accents, classic style, cozy and inviting, high detail, ultra-realistic",
    // 23
    "ancient library, high shelves filled with old books, detailed wood carvings, dusty and dim lighting, massive wooden tables, vintage globes, warm light filtering through tall windows, ultra-realistic, intricate details on book spines, nostalgic atmosphere, high resolution, serene and historical feel",

    // Fig. 7
    "A cat holding a sign that says hello world"

\end{lstlisting}
\caption{GPT-generated prompts used in the main manuscript.}
\label{lst:prompt}
\end{minipage}
\end{figure*}
\clearpage

\begin{figure*}[t]
\centering
\definecolor{backgroundcolor}{RGB}{250,250,250}
\begin{minipage}{\textwidth}
\lstset{
    language=xml,
    basicstyle=\tiny\ttfamily,
    backgroundcolor=\color{backgroundcolor},
    xleftmargin=0pt,
    breaklines=true,
    frame=single,
    showstringspaces=false,
    keepspaces=true
}
\begin{lstlisting}
    // Fig. 13, from top to bottom, from left to right

    // 1
    "a polar bear sitting on a floating iceberg, holding an umbrella while it rains colorful paint, the surrounding ocean reflecting the vibrant colors, ultra-detailed, photorealistic, ultra HD, 8K, surreal and artistic composition, bold contrasts, intricate reflections",
    // 2
    "lush green valley surrounded by towering cliffs, a winding river reflecting the blue sky, fluffy white clouds casting shadows, grazing deer in the distance, ultra-detailed, photorealistic, ultra HD, 8K, natural vibrancy, peaceful wilderness atmosphere, intricate water and vegetation textures",
    // 3
    "peaceful Chinese lake scene, a traditional pagoda on a small island, still water reflecting the structure, distant misty mountains, pink lotus flowers floating, warm morning light, ultra-detailed, photorealistic, ultra HD, 8K, serene atmosphere, traditional aesthetics, vibrant yet soft colors",
    // 4
    "owl in a dense forest at night, glowing yellow eyes, dark and mysterious atmosphere",
    // 5
    "a library where the books are glowing jellyfish floating mid-air, a young girl reaching out to touch one, shelves filled with ancient tomes, soft ambient lighting, ultra-detailed, photorealistic, ultra HD, 8K, whimsical and magical atmosphere, intricate textures",
    // 6
    "cyberpunk cityscape, glowing neon lights, futuristic skyscrapers, bustling streets, flying cars, nighttime setting, holographic advertisements, rain-soaked roads, ultra-detailed, cinematic lighting, ultra HD, 8K, vivid colors, dramatic atmosphere, intricate reflections, dystopian vibe",
    // 7
    "a futuristic robot tending a garden of glowing bioluminescent flowers, its metallic hands delicately handling the plants, a waterfall of stars in the background, ultra-detailed, photorealistic, ultra HD, 8K, ethereal lighting, blending nature and technology",
    // 8
    "alien desert landscape, multiple moons in the sky, strange rock formations, glowing plants, mysterious alien figures, science fiction style, ultra-detailed, cinematic lighting, ultra HD, 8K, vibrant colors, surreal ambiance, dramatic shadows, expansive vistas",
    // 9
    "autumn forest in golden hour, trees with vibrant red, orange, and yellow leaves, a narrow path covered in fallen foliage, sunlight casting warm hues, distant hills, ultra-detailed, photorealistic, ultra HD, 8K, rich colors, peaceful atmosphere, intricate details of leaves and bark",
    // 10
    "endless lavender fields at sunset, soft purple hues blending with golden sky, a small rustic farmhouse in the distance, rolling hills on the horizon, ultra-detailed, photorealistic, ultra HD, 8K, delicate lavender flowers, serene ambiance, atmospheric depth",
    // 11
    "cozy rural kitchen, wooden cabinets, fresh bread on the counter, sunlight streaming through lace curtains, ceramic jars and fresh herbs, rustic charm, warm tones, ultra-detailed, photorealistic, ultra HD, 8K, soft ambient light, intricate wood grain textures, peaceful atmosphere",
    // 12
    "portrait of a young girl with freckles, natural outdoor setting, sunlight filtering through leaves, soft focus background, vibrant hair and vivid eye color, ultra-detailed, photorealistic, ultra HD, 8K, delicate facial textures, bright and innocent atmosphere, warm golden tones"

    // Fig. 14, according to the top-left corner, from top to bottom, from left to right

    // 1
    "sunlit vineyard in late summer, rows of grapevines heavy with ripe fruit, rustic farmhouse in the distance, soft hills and clear sky, warm golden light, ultra-detailed, photorealistic, ultra HD, 8K, intricate grape and leaf textures, serene countryside atmosphere",
    // 2
    "ancient Chinese temple on a hill, red walls and golden roofs, surrounded by lush green bamboo forest, stone lanterns lining the path, soft golden hour light, ultra-detailed, photorealistic, ultra HD, 8K, traditional Chinese architecture, peaceful ambiance, intricate carvings and ornate designs",
    // 3
    "bustling fish market at sunrise, vibrant colors of fresh seafood, fishermen unloading crates, intricate details of fish scales and ice, ambient light, bustling atmosphere, ultra-detailed, photorealistic, ultra HD, 8K, atmospheric realism, sharp textures, lively dynamics",
    // 4
    "majestic dragon flying over a medieval castle, fiery sunset, rolling hills, dramatic clouds, fantasy style, ultra-detailed, painterly aesthetic, ultra HD, 8K, warm hues, glowing embers, intricate textures, golden hour lighting",
    // 5
    "futuristic laboratory interior, glowing screens, robotic arms, holographic displays, sleek design, science fiction style, ultra-detailed, ultra HD, 8K, cold lighting, metallic textures, high-tech ambiance, detailed equipment",
    // 6
    "a serene Chinese garden, a curved stone bridge over a lotus-filled pond, elegant pavilions with ornate designs, weeping willow trees, koi fish swimming, gentle sunlight, ultra-detailed, photorealistic, ultra HD, 8K, traditional landscape design, tranquil atmosphere, vibrant yet harmonious colors",
    // 7
    "ancient Greek temple on a hilltop, surrounded by lush gardens, golden hour, marble columns, intricate carvings, mythological figures, painterly style, ultra-detailed, ultra HD, 8K, warm lighting, serene atmosphere, historical accuracy",
    // 8
    "a chessboard floating in a cosmic void, pieces made of planets and stars, a human hand reaching out to make a move, ultra-detailed, photorealistic, ultra HD, 8K, cosmic and abstract design, vivid lighting, surreal and thought-provoking atmosphere",
    // 9
    "a vintage gramophone in the middle of a lush rainforest, vines wrapping around the horn, music notes visibly floating in the air, animals like parrots and monkeys curiously gathered, ultra-detailed, photorealistic, ultra HD, 8K, vibrant colors, magical and whimsical atmosphere, rich textures",
    // 10
    "a giant clock embedded in a mountain cliff, waterfalls flowing through the clock's gears, lush greenery surrounding the scene, ultra-detailed, photorealistic, ultra HD, 8K, timeless and surreal atmosphere, intricate mechanical details, dramatic lighting",
    // 11
    "sunset over a rocky coastline, waves crashing against jagged cliffs, vivid orange and purple hues in the sky, seabirds flying above, tide pools with reflections, ultra-detailed, photorealistic, ultra HD, 8K, dynamic motion, tranquil yet dramatic atmosphere, intricate rock and water textures",
    // 12
    "a gigantic hourglass buried in a desert, golden sand slowly flowing between the chambers, a group of explorers climbing the hourglass, a storm brewing in the background, ultra-detailed, photorealistic, ultra HD, 8K, dramatic lighting, surreal and adventurous ambiance",
    // 13
    "portrait of a wise old man with a long white beard, wearing traditional robes, holding a wooden staff, mountain landscape in the background, soft diffused light, ultra-detailed, photorealistic, ultra HD, 8K, deep wrinkles, serene expression, mystical and timeless atmosphere"
\end{lstlisting}
\caption{GPT-generated prompts used in the appendix.}
\label{lst:prompt-supp}
\end{minipage}
\end{figure*}
\clearpage

\end{document}